\documentclass[twocolumn]{autart}    

\usepackage{graphicx}          

\usepackage{amssymb}  
\usepackage{mathrsfs} 
\usepackage{graphicx,cite}
\usepackage{enumerate}
\usepackage{comment}
\usepackage{algorithm}
\usepackage{algpseudocode}
\usepackage{tcolorbox}
\usepackage{tabularx}
\usepackage{array}
\usepackage{dsfont}
\usepackage{tikz}
\usetikzlibrary{arrows.meta} 
\usepackage{amsmath}

\usepackage{setspace}
\newcommand{\remend}{\relax\ifmmode\else\unskip\hfill\fi\hbox{$\bullet$}}

\begin{document}

\begin{frontmatter}

\title{Provably Safe Reinforcement Learning for Stochastic Reach-Avoid Problems with Entropy Regularization\thanksref{footnoteinfo}} 

\thanks[footnoteinfo]{The work of the first and the second authors has been supported by the Independent Research Fund Denmark (DFF) under Project SAFEmig (Project Number 3105-00173B). The material in this paper was not presented at any conference.} 


\author[]{Abhijit Mazumdar\thanksref{cor}}\ead{abma@es.aau.dk},
\author[]{Rafal Wisniewski}\ead{raf@es.aau.dk},               
\author[]{Manuela  L. Bujorianu}\ead{lmbu@es.aau.dk}  
\thanks[cor]{Corresponding author.}
\address{Section of Automation $\&$ Control, Aalborg University, 9220 Aalborg East, Denmark}


          
\begin{keyword}                           
Safe reinforcement learning, constrained Markov decision processes, safety, entropy regularization.              
\end{keyword}                             

\begin{abstract}                          
We consider the problem of learning the optimal policy for Markov decision processes with safety constraints. We formulate the problem in a stochastic reach-avoid setup. Our goal is to design online reinforcement learning algorithms that ensure safety constraints with arbitrarily high probability during the learning phase. To this end, we first propose an algorithm based on the \textit{optimism in the face of uncertainty (OFU)} principle. Based on the first algorithm, we propose our main algorithm, which utilizes entropy regularization. We investigate the finite-sample analysis of both algorithms and derive their regret bounds. We demonstrate that the inclusion of entropy regularization improves the regret and significantly controls the episode-to-episode variability inherent in OFU-based safe RL algorithms.  
\end{abstract}

\end{frontmatter}

\section{Introduction}
Despite the tremendous successes of reinforcement learning (RL) across diverse domains, applying RL to safety-critical domains, such as healthcare, autonomous vehicles, and power grids, remains challenging because standard RL methods lack mechanisms to ensure safety, both during training and execution. To address this, safe reinforcement learning builds upon the traditional RL framework by explicitly enforcing safety constraints and guiding agents away from actions that could result in unsafe outcomes. Safe reinforcement learning is typically formulated within a constrained Markov decision process (CMDP) setup \cite{altman1999constrained}.

As practical systems are inevitably subject to stochastic disturbances and uncertainty, the safety notion considered in this paper is probabilistic in nature. We adopt the concept of \(p\)-safety, first introduced in \cite{wisniewski2017stochastic} and further developed in \cite{wisniewski2020p,BuWiBou2020,wisniewski2021safety}. In the context of Markov decision processes (MDPs), model-free methods for assessing \(p\)-safety have been studied in \cite{wisniewski2023probabilistic,mazumdar2023online,mazumdar2024online}. 
 \\
 \\
 \textbf{\textit{Related literature:}} A variety of safe RL methods have been developed in the recent past.  
 \par
 \par
\textit{Linear Programming Approach:}
CMDPs form the theoretical backbone of safe RL by extending standard MDPs to account for constraints on expected cumulative costs. The foundational work \cite{altman1999constrained} introduces CMDPs using linear programming (LP) and Lagrangian duality. Building on the works presented in \cite{altman1999constrained}, \cite{efroni2020exploration} proposes a LP-based learning algorithms that achieve sublinear regret in online settings while respecting constraints. More recently, \cite{singh2022learning} develops algorithms that carefully balance exploration and constraint satisfaction. In \cite{bura2022dope}, another algorithm is developed that ensures safety throughout the learning phase and provides a better theoretical regret bound. However, all of the aforementioned works consider either a deterministic finite horizon or an infinite horizon.
\par \textit{Lagrangian and Primal-Dual Approach:}
Another key direction in safe RL is the use of Lagrangian-based methods, which convert constrained optimization problems into unconstrained ones by introducing penalty terms. Reference \cite{achiam2017constrained} presents constrained policy optimization (CPO), a practical algorithm that adapts trust-region policy updates to enforce constraint satisfaction. Further, \cite{wei2022provably,wei2023provably} introduce Q-learning-based methods that combine improved convergence properties with enhanced empirical performance, particularly in episodic CMDPs.
\par \textit{Safety Filters and Shielding Mechanisms:}
Safety filters provide a mechanism for ensuring real-time safety by intervening when an agent proposes unsafe actions. These approaches often rely on models of the environment to verify whether an action will violate safety constraints. References \cite{alshiekh2017safe, jansen2020safe} introduce shielding via formal methods to prevent unsafe transitions. Furthermore, \cite{bastani2021safe} introduces probabilistic shields that can handle uncertainty during both the training and deployment phases.
\par \textit{Policy Gradient and Actor-Critic Methods:}
Gradient-based methods have also been adapted to the safe learning setting by modifying their objectives to incorporate constraint satisfaction. A convergent actor-critic algorithm for CMDPs with provable guarantees is introduced in \cite{yu2019convergent}. In \cite{ding2020natural}, a natural policy gradient approach is designed with a non-asymptotic convergence guarantee. In \cite{papini2022smoothing}, the challenge of sample efficiency in constrained environments is addressed by proposing a smoothing technique that balances learning performance with safety. 
\par The use of entropy regularization in RL receives significant attention due to its ability to promote exploration, stabilize learning, and induce smoother policy updates. We briefly review representative works below.

\textit{Entropy-regularized RL:}
 Maximum-entropy principles are formalized in \cite{ziebart2008maximum} to derive stochastic policies that balance reward maximization and exploration. Entropy regularization is now a standard component of policy-gradient methods, where it is commonly employed to prevent premature policy collapse and to improve optimization stability, as illustrated by trust-region and proximal policy optimization algorithms \cite{schulman2015trust,schulman2017proximal}. The Soft Actor-Critic (SAC) framework explicitly incorporates entropy maximization into the RL objective and demonstrates strong empirical performance and robustness in continuous-control tasks \cite{haarnoja2018soft}. In \cite{neu2017unified}, entropy-regularized reinforcement learning methods are shown to cast many state-of-the-art algorithms as approximate instances of Mirror Descent or Dual Averaging. However, these entropy-regularized approaches are developed in unconstrained settings and do not provide formal safety guarantees. 
\\
\\
\textbf{\textit{Our contributions:}} We consider the problem of learning an optimal policy for Markov decision processes (MDPs) subject to a safety constraint. Adopting a probabilistic notion of safety, namely \(p\)-safety \cite{wisniewski2021safety}, we formulate the problem within a stochastic reach--avoid framework. We assume that the state space of the MDP is partitioned into three disjoint subsets: a \emph{target or goal set} comprising all goal states, an \emph{unsafe set} comprising all unsafe states, and a \emph{living set} consisting of all remaining states that are neither goal nor unsafe.

To formalize this setup, we employ a CMDP formulation \cite{altman1991adaptive}. We assume that the target set is terminal, meaning that the process terminates whenever the MDP reaches a target state, while the unsafe set and the living set are non-terminal. In contrast, our earlier work \cite{mazumdar2024safe} assumes that the unsafe set is also terminal to reduce computational complexity. The setup considered in \cite{mazumdar2024safe} precludes transitions from the unsafe set to other states, thereby simplifying the underlying optimization problem. In this work, however, we adopt a more general and practically motivated modeling framework in which only the target set is terminal, while transitions from the unsafe set to the living set or directly to the goal set are permitted. This formulation captures a broad class of real-world systems in which safety violations are not necessarily absorbing events, but rather correspond to transient or recoverable conditions that incur increased risk. For instance, an autonomous robot may temporarily enter unsafe regions before corrective actions are applied, and clinical decision-making processes may involve adverse but reversible outcomes. By allowing non-absorbing unsafe states, the model more accurately reflects the underlying system dynamics and enables the development of learning algorithms that explicitly trade off performance and probabilistic safety.

The main contributions of this paper are as follows.

\begin{enumerate}[i)]
    \item We first propose a reinforcement learning (RL) algorithm, termed \emph{\(p\)-Safe RL (pSRL)}, that is based on the optimism under uncertainty (OFU) principle. The proposed algorithm constitutes a non-trivial extension of our earlier work \cite{mazumdar2024safe}. Unlike \cite{mazumdar2024safe}, which requires the existence of a deterministically safe action (i.e., an action that is safe with probability one) to construct a baseline policy, the present work relaxes this requirement by allowing the baseline policy to rely on probabilistically safe actions. Furthermore, we provide a comprehensive theoretical analysis of the proposed algorithm, including the derivation of a formal regret bound, which is not addressed in \cite{mazumdar2024safe}.

    \item We then introduce a novel algorithm, termed \emph{Entropy-Regularized \(p\)-safe RL (ER-pSRL)}, which augments the OFU-based pSRL algorithm with an entropy regularization term. We show that this modification leads to improved regret performance, both theoretically and empirically. To the best of our knowledge, ER-pSRL is the first provably safe reinforcement learning algorithm to incorporate entropy regularization and to achieve regret bounds that improve upon existing state-of-the-art results.
 
    \item We demonstrate that entropy regularization substantially reduces episode-to-episode variability in the per-episode regret. In contrast, the OFU-based pSRL algorithm exhibits highly non-smooth policy updates, leading to large fluctuations in per-episode regret across episodes. 
    Moreover, the ER-pSRL algorithm achieves a lower \emph{cumulative objective regret} compared to the pSRL algorithm.

    \item Finally, we show that incorporating prior knowledge of a subset of the state space, termed the \textit{proxy set}, further accelerates learning and leads to improved performance compared to learning without such structural information.
\end{enumerate}

The remainder of the paper is structured as follows. In Section \ref{background}, we establish the necessary background and introduce the learning problem. In Section \ref{p-safe_algo}, we present our first algorithm, namely the \emph{p-safe RL (pSRL)} algorithm, along with its theoretical guarantees. In Section \ref{entropy_algo}, we develop our second algorithm, the \emph{entropy-regularized $p$-safe RL (ER-pSRL)} algorithm. We then analyze the theoretical results by using a numerical example in Section \ref{simulation_section}. Finally, we conclude the paper in Section \ref{sec_conclusion} and discuss how this work can be extended in various direction. Most of the proofs of the results presented in the paper are provided in the Appendix Section.\\

\section{Background and Problem Formulation} \label{background}

We consider a Markov decision process (MDP) $M=\left(\mathcal{X}, \mathcal{A}, P, c \right)$ with a finite set of states, $\mathcal{X}$, and a finite set of actions, $\mathcal{A}$. The state set $\mathcal{X}$ is composed of three subsets: the set of all \textit{goal or target} states $E$, the set of \textit{forbidden or unsafe} states $U$, and the \textit{living set} $H := \mathcal{X} \setminus (E \cup U)$. In our setup, both the living set $H$ and the unsafe set $U$ are transient, and the goal set $E$ is terminal, meaning that the process ends upon reaching set $E$. We use uppercase letters $X_t$ and $A_t$ to represent the random state and action at time $t$, while lowercase letters $x_t$ and $a_t$ denote their corresponding deterministic values or realizations. We consider a sample space $\Omega$ of the form $\omega = (x_0, a_0, x_1, a_1, \ldots) \in (\mathcal{X} \times \mathcal{A})^{\infty}$, where $x_i \in \mathcal{X}$ and $a_i \in \mathcal{A}$. The sample space $\Omega$ is equipped with a $\sigma$-algebra $\mathcal{F}$, generated by the coordinate map $(X_t \times A_t)(\omega) = (x_t,a_t)$. A policy is denoted by $\pi(a \mid x)$, where $\pi : \mathcal{X} \to \Delta(\mathcal{A})$ and $\Delta(\mathcal{A})$ denotes the probability simplex over the action set $\mathcal{A}$. The transition dynamics are assumed to be time-invariant and are described by the transition probability function
$P(x,a,y) := \mathds{P}\!\left[ X_{t+1} = y \mid X_t = x,\, A_t = a \right],
\quad (x,a,y) \in \mathcal{X} \times \mathcal{A} \times \mathcal{X}$. For notational convenience, we write $P(x,a,\cdot) := \{ P(x,a,y) \mid y \in \mathcal{X} \}$ and $\qquad P := \{ P(x,a,\cdot) \mid (x,a) \in \mathcal{X} \times \mathcal{A} \}$. Given two transition probability functions $P$ and $P'$, we define their point-wise difference as $(P - P')(x,a,y) := P(x,a,y) - P'(x,a,y)$. When action $A_t = a\in \mathcal{A}$ is applied in state $X_t = x \in \mathcal{X}$, the agent incurs a cost $c(x,a)$. The total number of episodes is denoted by $K$. For a fixed policy $\pi$ and transition kernel $P$, we denote by $\mathds{P}_{\pi}^{x}[\,\cdot\,;P]$ the probability measure over trajectories induced by starting from state $x$. The corresponding expectation is denoted by $\mathds{E}_{\pi}^{x}[\,\cdot\,;P]$. We use $\tilde{\mathcal{O}}(\cdot)$ to denote asymptotic order up to polylogarithmic factors, and the notation $\overset{<}{\sim}$ to indicate an inequality that holds up to universal constants or polylogarithmic terms. For $a,b \in \mathds{R}$, we write $a \wedge b := \min\{a,b\}$ and $a \vee b := \max\{a,b\}$. Finally, for functions $f,g : \mathcal{X} \to \mathds{R}$, the inner product $\langle f, g \rangle$ is defined as $\langle f, g \rangle := \sum_{y \in \mathcal{X}} f(y) g(y)$.

We consider a probabilistic notion of safety, referred to as \emph{$p$-safety} \cite{wisniewski2023probabilistic}. This notion captures the probability that the system visits an unsafe state before reaching a goal state. Accordingly, we define a \emph{safety function} that quantifies the probability of entering the set of forbidden states prior to reaching the goal set. For an MDP $M$ with transition probability kernel $P$ and a policy $\pi$, the safety function $S^P_{\pi}(x_0)$ is defined for any initial state $x_0$ as follows:
\small
\begin{equation*}
    S_{\pi}^P(x_0) := \mathds{P}^{x_0}_{\pi}[\tau_U < \tau_{E} ; P],
    \label{safety_DP}
\end{equation*}
\normalsize
where $\tau_A$ is the first hitting time of set $A$.\\
 Following \cite{wisniewski2023probabilistic}, we can express the safety function as:
 \small
\begin{equation}
    S^P_{\pi}(x_0) = \mathds{E}_{\pi}^{x_0} \left[ \sum_{t = 0}^{\tau-1} \kappa(X_t, A_t) ; P \right],
    \label{safety_fn_exprsn}
\end{equation}
\normalsize
where $\tau := \tau_{E \cup U}$, and $\kappa(x,a) := \sum_{y \in U} P({x,a,y})$ for all $x \in \mathcal{X} \setminus (E\cup U)$. \\
Another recursive expression of the safety function can be derived \cite{mazumdar2024safe}:
\small
\begin{equation*}
        S^P_{\pi}(x_0) = \mathds{E}_{\pi} \Big[\kappa(x_0,a_0) + \sum_{y} P(x_0,a_0,y) S^P_{\pi}(y) \big| a_0 = \pi(x_0) \Big].
    \end{equation*}
\normalsize
\par For $t\leq \tau_E-1$, we define a sequence $\{c_t\}$ with $c_t:=c(X_t,A_t)$ and for all $t\geq\tau_E$, $c_t=0$. Similarly for $t\leq \tau-1$, we define another process $\{\kappa_t\}$ with $\kappa_t:= \kappa(X_t,A_t)$ and for all $t\geq \tau$, $\kappa_t:= 0$.
\begin{defn}[$p$-safety]
   Consider a given safety parameter $p\in (0,1)$, an initial state $X_0={x}_0 \in H$ and a policy $\pi$. If $S^P_{\pi}({x}_0)\leq p$, then the initial state ${x}_0$ (the policy $\pi$) is called a $p$\textit{-safe} or simply \textit{safe initial state} (\textit{safe policy}). 
\end{defn}
In our earlier work \cite{mazumdar2024safe}, the notion of a \emph{safe action} is restricted to deterministic actions. In the present work, we relax this assumption and demonstrate that a probabilistically safe action, as defined below, is sufficient.

\begin{defn}[Safe action] 
   Let $p \in (0,1)$ and let $h$ be a scalar satisfying $0 \leq h < p$. An action $a(x) \in \mathcal{A}$ is said to be {safe} if, when action $a(x)$ is applied at state $x$, the probability of transitioning to the unsafe set $U$ satisfies $\sum_{y \in U} P(x,a(x),y) \leq h .$

\end{defn}
We also introduce a subset of the living set $H$, referred to as the \textit{proxy set}, consisting of states that are probabilistically ``close'' to the unsafe set $U$. As shown in the numerical section, incorporating the proxy set enables more efficient learning and leads to improved policy performance.
\begin{defn}[Proxy Set] \cite{mazumdar2023online}\label{Def.ProxySet}
Let $U' \subset H$. The set $U'$ is said to be a \textit{proxy set} if the following conditions hold:
\begin{itemize}
    \item For any policy $\pi \in \Delta(\mathcal{A})$ and any initial state $x_0 \in H$, the stopping time $\tau_{U'}$ occurs strictly before $\tau_U$ almost surely, i.e., $\tau_{U'} < \tau_U$ almost surely.
    \item From any state $x \in U'$, the unsafe set $U$ is reached in a single transition with probability one, i.e.:
    \small
    \[\sum_{a \in \mathcal{A}} \sum_{y \in U} P(x,a,y) = 1.\]
    \normalsize
\end{itemize}
\end{defn}
If a safe baseline policy is enforced only on the proxy states while an arbitrary (possibly exploratory) policy is applied elsewhere, the overall safety of the system remains guaranteed \cite{mazumdar2024online}. Hence, the proxy set enables enhanced exploration by allowing less conservative actions without compromising safety. This work is, however, not restricted to the prior knowledge of the proxy set. In scenarios where the proxy set is unknown or cannot be identified apriori, we adopt a conservative assumption by treating all states as proxy states.

\textbf{\textit{The learning problem:}}\\
\\ We strive to find a policy $\pi$ that minimizes the cumulative cost starting from an initial state until the MDP reaches the goal set $E$, i.e., solve for the policy 
\begin{equation}
    \begin{split}
        &  \underset{\pi}{{argmin }} \ V_{\pi}^P(c,x_0)  \\
     & {subject \ to } \ S^P_{\pi}({x}_0) \leq p
    \end{split}
    \label{safety_prob}
\end{equation}
where $V^P_{\pi}(c,x)$ is the cumulative cost function to be minimized and is defined as
\begin{equation*}
     V^{P}_{\pi}(c,x) := \mathds{E}^{{x}}_{\pi} \left[ \sum_{t=0}^{\tau_E -1} c_t ; P \right].
\end{equation*}
\normalsize
\par
Throughout the paper, we have the following standing assumptions:  \\
\begin{assum} \label{assum_access2}
    For all deterministic policies, the goal set $E$ is accessible from all living states $H$ and from all unsafe states in $U$, meaning that starting from any state in the set $H\cup E$, there is a nonzero probability to visit the goal set $E$.
\end{assum}
\begin{assum}\label{assum_transient}
 For all deterministic policies, all states in the living set $H$, and the unsafe set $U$ are \textit{transient}, meaning that the process leaves the set $H\cup U$ in finite time-steps almost surely. 
 \end{assum}
   Assumption \ref{assum_access2} and Assumption \ref{assum_transient} are required to have a finite stopping time for all policies (deterministic or stochastic), thus making the problem well-defined \cite{wisniewski2023probabilistic}.

 Due to Assumption \ref{assum_transient}, the hitting time of set $E$ is finite almost surely. Further, set $E$ is accessible from set $H\cup U$ for any policy $\pi$. Hence, the stopping time $\tau_E$ is upper bounded by a positive real $T_{max}$ almost surely, i.e., $\mathds{P}^{x_0}_{\pi}[\tau_E \leq T_{max} ; P]=1$. 
\begin{assum} \label{cost_assum}
Without loss of generality, we assume that the cost function $c(x,a)$ is uniformly bounded as $|c(x,a)| \leq 1$ for all $(x,a) \in (H \cup U) \times \mathcal{A}$. Moreover, since the target set $E$ is terminal, we can assume without further restriction that $c(x,a) = 0$ for all $(x,a) \in E \times \mathcal{A}$.
\end{assum}
 The following assumption guarantees the existence of a safe baseline policy for the proposed algorithms.
\begin{assum}\label{safe_action_assum}
For every state $x \in U'$, there exists at least one safe action $a^s(x)$ satisfying $h \le \frac{p}{T_{\max}}$.
\end{assum}

In the next section, we present our first algorithm which is based on the OFU principle.
\section{p-Safe Reinforcement Learning (pSRL) Algorithm} \label{p-safe_algo}


In this section, we first develop an algorithm to learn the optimal policy for the learning problem \eqref{safety_prob}. 


 If the transition probabilities are known, then as shown in \cite{wisniewski2023probabilistic}, we could solve problem \eqref{safety_prob} by converting it into a linear program. To this end, we define the \textit{state-action occupation measures} or simply \textit{occupation measures} $\gamma(x,a)$, $\beta(x,a)$, 
\begin{equation}
\begin{split}
 &   \gamma(x,a) = \sum_{t=0}^{\infty} \mathds{P}^{x_0}_{\pi}[X_t=x,A_t=a,t < \tau ; P]    \\ 
 &  \beta(x,a) = \sum_{t=0}^{\infty} \mathds{P}^{x_0}_{\pi}[X_t=x,A_t=a,\tau \leq t < \tau_E ; P] \\
& \text{recall that }  \tau:= \tau_{E\cup U}.
\end{split}
\label{occu_measure}
\end{equation}
 We also define the \emph{total state-action occupation measures} $\xi(x,a)$ as follows:
 \begin{equation*}
       \xi(x,a) = \gamma(x,a) + \beta(x,a).
 \end{equation*}
As shown in \cite{wisniewski2023probabilistic}, we can represent the cost function $V_{\pi}^P(c)$ as follows:
\begin{equation*}
\begin{split}
    & V_{\pi}^P(c) = \sum_{(x,a)\in \mathcal{X}\times \mathcal{A}} {\xi}(x,a) c(x,a).
\end{split}
\end{equation*}

Using Proposition $2$ in \cite{wisniewski2023probabilistic}, we convert Problem \eqref{safety_prob} to the following linear programming: 
\small
\begin{subequations} \label{known_LP}
    \begin{equation}
    \begin{split}
      & \min_{\gamma \geq 0, \beta \geq 0}  \sum_{(x,a) \in \mathcal{X} \times \mathcal{A}} ({\gamma}(x,a) + {\beta}(x,a)) c(x,a) \\  
    \end{split}
    \end{equation}
    \begin{equation}
    \begin{split}
     & \mathbf{s.t. }\\
     & {\gamma}(x, a) = 0, \hbox{ for } (x,a) \in (U \cup E)\times \mathcal{A} \\
      & {\beta}(x, a)=0, \hbox{ for } (x,a) \in E \times \mathcal{A}\\   
    \end{split}
    \end{equation}
     \begin{equation}
    \begin{split}
     & \forall \ y \in H: \\
      & \delta_{x_0}(y) + \sum_{(x,a) \in H \times \mathcal{A}} {\gamma}(x, a) (P(x,a,y) - \delta_y(x)) = 0,
    \end{split}
    \end{equation}
    \begin{equation}
    \begin{split}
      & \sum_{(x,a) \in (H\cup U)  \times \mathcal{A}}  {\beta}(x, a) (P(x,a,y) - \delta_{y}(x)) = 0 , 
    \end{split}
    \end{equation}
     \begin{equation}
    \begin{split}
     & \forall \ y\in U: \\
      &  \sum_{(x,a) \in H \times \mathcal{A}}  \left({\gamma}(x, a) + {\beta}(x, a)\right) (P(x,a,y) - \delta_y(x)) \\
      &+ \sum_{(x,a) \in U \times \mathcal{A} } {\beta}(x, a) (P(x,a,y) - \delta_y(x))) = 0, 
    \end{split}
    \end{equation}
     \begin{equation}
    \begin{split}
      & \sum_{(x,a) \in H \times \mathcal{A}} {\gamma}(x,a) \kappa(x,a) \leq p. 
    \end{split}
    \end{equation}
\end{subequations}
\normalsize
The optimal policy for all $x\in H\cup U$ that solves problem \eqref{safety_prob} is given by:
\begin{align}\label{PolicyGenerationConstraints}
  \pi^{*}(a|x)  = \frac{{\gamma}^{*}(x,a) + \beta^{*}(x,a)}{\sum_{a \in \mathcal{A}} \left({\gamma}^{*}(x,a) + \beta^{*}(x,a)\right)},  
\end{align}
where, $\gamma^{*}(x,a)$ and $\beta^{*}(x,a)$, $\forall(x,a)\in H \cup U \times \mathcal{A}$, are the solution of the above LP \eqref{known_LP}. 
\par Since the transition probability kernel $P$ is not known, the above LP cannot be used directly. However, one can estimate the transition probabilities, e.g.  by using the empirical probability denoted by $\hat{P}(x,a,y)$, and then solve LP \eqref{known_LP}. While this seems quite intuitive and straightforward, this naive approach faces the following substantial difficulties:
\begin{enumerate}[i)]
\item There is no guarantee that with empirical probability $\hat{P}$ LP \eqref{known_LP} is feasible. 
\item Even if it is feasible, replacing the true probability $P$ with $\hat{P}$ could result in finding policies that are unsafe for the original problem.
\item Exploration is limited.
\end{enumerate}

We design our algorithm with these bottlenecks in mind. Since the stopping time is almost surely finite, the resulting learning procedure is episodic. Each episode $k \in \{1,2,3,\ldots\}$ starts from a fixed initial state $x_0$ and terminates upon reaching a goal state. Let $X_{k,t}$ and $A_{k,t}$ denote the random state and action at time step $t$ during episode $k$. When the discussion pertains to a single episode, we omit the episode index and simply write $X_t$ and $A_t$. At the beginning of each episode $k \geq 2$, we compute the empirical transition distribution as follows:

\scriptsize
\begin{equation}
\begin{split}
& \hat{P}_{k}(x,a,y):= \frac{\sum_{k'=1}^{k-1} \sum_{t=0}^{\tau_E-1} \mathds{1} \left({X}_{k',t} =x, {A}_{k',t} = a, {X}_{k'+1,t} =y \right)}{N_{k}(x,a)\vee 1}
\end{split}
\label{computed_prob}
\end{equation}
\normalsize
where $N_{k}(x,a)$ is the given by 
\small
     \begin{equation*}
     \begin{split}
        & {N}_{k}(x,a): = \sum_{k'=1}^{k-1} \sum_{t=0}^{\tau_E-1} \mathds{1} \left({X}_{k',t} =x, {A}_{k',t} = a \right).
    \end{split}
     \end{equation*}
     For episode $k=1$, we pick any random distribution for $\hat{P}_k(x,a,y)$.

At each episode $k$, we solve the following optimization, which is a modification of LP \eqref{known_LP}.
\small
\begin{equation} \label{ofu_pol}
    \begin{split}
        &\underset{\tilde{P} \in \mathcal{M}_k,\hat{\gamma}\geq 0, \hat{\beta}\geq 0}{\text{min}}  \ \sum_{(x,a)\in \mathcal{X}\times \mathcal{A}} (\hat{\gamma}(x,a)+\hat{\beta}(x,a)) \tilde{c}_k(x,a) \\
    \end{split}
\end{equation}
\begin{subequations} \label{ofu_pol_feas}
\begin{equation}
    \begin{split}
    & \mathbf{s.t. } \\
    & \hat{\gamma}(x, a) = 0, \hbox{ for } (x,a) \in (U \cup E)\times \mathcal{A} \\
      & \hat{\beta}(x, a)=0, \hbox{ for } (x,a) \in E \times \mathcal{A}\\   
    \end{split}
    \end{equation}
     \begin{equation}
    \begin{split}
     & \forall \ y \in H: \\
      & \delta_{x_0}(y) + \sum_{(x,a) \in H \times \mathcal{A}} \hat{\gamma}(x, a) (\tilde{P}(x,a,y) - \delta_y(x)) = 0,
    \end{split}
    \end{equation}
    \begin{equation}
    \begin{split}
      & \sum_{(x,a) \in (H\cup U)  \times \mathcal{A}}  \hat{\beta}(x, a) (\tilde{P}(x,a,y) - \delta_{y}(x)) = 0 , 
    \end{split}
    \end{equation}
     \begin{equation}
    \begin{split}
     & \forall \ y\in U: \\
      &  \sum_{(x,a) \in H \times \mathcal{A}}  \left(\hat{\gamma}(x, a) + \hat{\beta}(x, a)\right) (\tilde{P}(x,a,y) - \delta_y(x)) \\
      &+ \sum_{(x,a) \in U \times \mathcal{A} } \hat{\beta}(x, a) (\tilde{P}(x,a,y) - \delta_y(x))) = 0, 
    \end{split}
    \end{equation}
     \begin{equation}
    \begin{split}
      & \sum_{(x,a) \in H \times \mathcal{A}} \hat{\gamma}(x,a)  (\hat{\kappa}(x,a)+3\hat{\epsilon}_k(x,a)) \leq p, 
    \end{split}
    \end{equation}
\end{subequations}
\normalsize
where, 
\small
\begin{equation}\label{para_ofu}
    \begin{split}
        & \tilde{c}_k(x,a)= c(x,a) - \frac{4T_{max}}{p-p^s} \hat{\epsilon}_k(x,a),\\
          & p^s  = S^P_{\pi^s}(x_0), \\
        &\mathcal{M}_k = \Big\{ \mathcal{Q}_k(x,a), \ \forall (x,a)\in (H\cup U)\times \mathcal{A} \Big\}. \\
       &\mathcal{Q}_k(x,a) = \Big\{ \tilde{P}(x,a,.)\in \Delta(\mathcal{X}) \;\Big|\;  \\
        & \hspace{1cm} | \tilde{P}(x,a,y) - \hat{P}_k(x,a,y)| \leq \epsilon_k(x,a,y), \ \ \; \forall y \in \mathcal{X} \Big\}. \\ 
    \end{split}
\end{equation}
We choose $\epsilon_k(x,a,y)$ according to the empirical Bernstein inequality \cite{maurer2009empirical}:
\begin{equation}
    \begin{split}
        & \epsilon_k(x,a,y)\\
       &= \sqrt{\frac{4\hat{P}_k(x,a,y)(1-\hat{P}_k(x,a,y))L}{N_k(x,a)\vee 1}}  + \frac{14L}{3(N_k(x,a)\vee 1)},  \\
       & \text{where } L = \text{ log}\big(\frac{2|\mathcal{X}||\mathcal{A}|K}{\delta}\big), \\
       & \text{further, } \hat{\epsilon}_k(x,a) = \sum_{y\in \mathcal{X}} \epsilon_k(x,a,y). 
    \end{split}
    \label{eps_expression}
\end{equation}
 \normalsize 
 The optimal policy is computed as:
 \begin{align*}
   {\pi}_k(a|x)=\frac{\hat{\gamma}^*(x,a) + \hat{\beta}^*(x,a)}{{\sum_{a\in \mathcal{A}}}\left( \hat{\gamma}^*(x,a) + \hat{\beta}^*(x,a)\right)},  
 \end{align*}
  where $\hat{\gamma}^*(x,a), \hat{\beta}^*(x,a)$ are the solution of optimization \eqref{ofu_pol}-\eqref{ofu_pol_feas}.

 The computed policy ${\pi}_k(a|x)$ is kept fixed till the end of the episode $k$. Optimization \eqref{ofu_pol}-\eqref{ofu_pol_feas} is based on the \textit{optimization in the face of uncertainty (OFU)} principle \cite{efroni2020exploration}. By introducing $\tilde{P}\in \mathcal{M}_k$ as the additional decision variable, optimization \eqref{ofu_pol} not only increases the chances of feasibility but also results in better regret bound \cite{efroni2020exploration}. However, as a result of this, optimization \eqref{ofu_pol} now becomes complex to solve as $\tilde{P}$ is also a decision variable. The optimal value of the transition probability $\tilde{P}$ that maximizes optimization \eqref{ofu_pol}-\eqref{ofu_pol_feas} is called the \emph{optimistic transition probability} and denoted by $\tilde{P}_k$.

Fortunately, by introducing \textit{state-action-state occupation measures} $\hat{h}(x,a,y):= \hat{\gamma}(x,a)\tilde{P}(x,a,y)$ and $\hat{g}(x,a,y):= \hat{\beta}(x,a)\tilde{P}(x,a,y)$, we can convert optimization \eqref{ofu_pol} into an \textit{extended linear programming} as follows:      
 \scriptsize
 \begin{equation}
     \begin{split}
         &  \underset{\hat{h}\geq 0, \hat{g}\geq 0}{\text{min}} \sum_{(x,a,y)\in \mathcal{X}\times \mathcal{A} \times \mathcal{X}} (\hat{h}(x,a,y)+\hat{g}(x,a,y)) \tilde{c}_k(x,a)
     \end{split}
     \label{ofu_ext_lp}
 \end{equation}
 \begin{subequations} \label{ofu_ext_lp_constraints}
 \begin{equation}
  \begin{split}
     & \mathbf{s.t. } \\
     & \hat{h}(x, a,y) = 0, \text{ for } (x,a,y) \in (U \cup E)\times \mathcal{A} \times \mathcal{X} \label{eq:one}
     \end{split}
 \end{equation}
 \begin{equation}
   \hat{g}(x, a, y)=0, \hbox{ for } (x,a,y) \in E \times \mathcal{A}\times \mathcal{X}    
 \end{equation}
 \begin{equation}
     \begin{split}
         & \text{for } y\in H:  \\
 & \delta_{x_0}(y) + \sum_{(x,a) \in H \times \mathcal{A}} \hat{h}(x, a, y) - \sum_{(z,a) \in \mathcal{X} \times \mathcal{A}} \hat{h}(y, a, z) = 0, \\
     \end{split}
 \end{equation}
\begin{equation}
    \begin{split}
         & \sum_{(x,a) \in (H\cup U)  \times \mathcal{A}}  \hat{g}(x, a,y)  - \sum_{(z,a) \in \mathcal{X} \times \mathcal{A}}  \hat{g}(y, a, z) = 0 , \\
    \end{split}
\end{equation}
\begin{flalign}
\begin{split}
 \text{for } y\in U&:  \\  
& \sum_{(x,a) \in H \times \mathcal{A}}  \big(\hat{h}(x, a, y) + \hat{g}(x, a, y) \big)  \\
      &+ \sum_{(x,a) \in U \times \mathcal{A} } \hat{g}(x, a ,y) - \sum_{(z,a) \in \mathcal{X} \times \mathcal{A} } \hat{g}(y, a ,z) = 0,
      \end{split}
\end{flalign}
\begin{equation}
    \begin{split}
      \forall (x,&a,y)\in (H\cup U)\times \mathcal{A}\times \mathcal{X}:  \\
     & \hat{h}(x,a,y) - (\hat{P}_k(x,a,y)+\epsilon_k(x,a,y))\sum_{z\in \mathcal{X}} \hat{h}(x,a,z) \leq 0,
    \end{split}
\end{equation}
\begin{equation}
    \begin{split}
         & -\hat{h}(x,a,y) + (\hat{P}_k(x,a,y)-\epsilon_k(x,a,y))\sum_{z\in \mathcal{X}} \hat{h}(x,a,z) \leq 0,
    \end{split}
\end{equation}
\begin{equation}
    \begin{split}
       & \hat{g}(x,a,y) - (\hat{P}_k(x,a,y)+\epsilon_k(x,a,y))\sum_{z\in \mathcal{X}} \hat{g}(x,a,z) \leq 0,  
    \end{split}
\end{equation}
\begin{equation}
    \begin{split}
      & -\hat{g}(x,a,y) + (\hat{P}_k(x,a,y)-\epsilon_k(x,a,y))\sum_{z\in \mathcal{X}} \hat{g}(x,a,z) \leq 0,  
    \end{split}
\end{equation}
\begin{equation}
    \begin{split}
          & \sum_{(x,a,y) \in H \times \mathcal{A}\times \mathcal{X}} \hat{h}(x,a,y) (\hat{\kappa}(x,a)+3\hat{\epsilon}_k(x,a)) \leq p.  
    \end{split}
\end{equation}
\end{subequations}
 \normalsize
 The policy is computed as:
 \begin{equation}
 {\pi}_k(a|x)=\frac{\sum_{y\in \mathcal{X}}\left(\hat{h}^*(x,a,y)+\hat{g}^*(x,a,y)\right)}{\sum_{(a',y)\in \mathcal{A}\times \mathcal{X}}\left(\hat{h}^*(x,a',y)+\hat{g}^*(x,a',y)\right)},
 \label{estimate_pol1}
\end{equation} 
where, $\hat{h}^*(x,a,y)$ and $\hat{g}^*(x,a,y)$ are the solution of the extended LP \eqref{ofu_ext_lp}.
\par If at any episode the extended LP \eqref{ofu_ext_lp} is not feasible, we apply a safe baseline policy. Once at an episode $k$, LP \eqref{ofu_ext_lp} is feasible, we demonstrate that the policy ${\pi}_k$ obtained using \eqref{estimate_pol1} is a safe policy with an arbitrarily high probability. The pseudo-code for the pSRL algorithm is shown in Algorithm~1. 

\small
\begin{algorithm} \label{safe_algo}
  \caption{: pSRL Algorithm}
  \begin{algorithmic}[1]
    \State \textbf{Input:} Safety parameter $p\in (0,1)$, initial state $x_0$, confidence parameter $\delta \in (0,1)$, one safe action $a^{s}(x')$ for each proxy state, cost $c(x,a)$ for all $(x,a)\in \mathcal{X} \times \mathcal{A}$
    \State \textbf{Initialize:} Select $N_{1}(x,a)=0$, $N_{1}(x,a,y)=0$ for all $(x,a,y)\in H\times \mathcal{A}\times \mathcal{X}$
    \For  {Episodes ($k=1,2,...,K$)}
    \State Set $X_0 \leftarrow x_0$
    \State Solve LP \eqref{ofu_ext_lp}-\eqref{ofu_ext_lp_constraints} using the estimate $\hat{P}_k(x,a,y)$ from \eqref{computed_prob}
    \If{ LP \eqref{ofu_ext_lp}-\eqref{ofu_ext_lp_constraints} is feasible}
    \State Obtain the policy ${\pi}_{k}$ using \eqref{estimate_pol1}
    \Else
    \State Set ${\pi}_{k} \leftarrow \pi^s$ 
    \EndIf
  \For  { ($t=0,1,2,...$)}
    \If  { $X_t$ is not a terminal state}
    \State  Apply action $A_t$ according to the policy ${\pi}_{k}$
    \State Observe the next state $X_{t+1}$  
      \Else 
     \State Terminate the loop
    \EndIf
     \EndFor
     \EndFor
  \end{algorithmic}
\end{algorithm}
\normalsize
\par
We now derive a safe baseline policy using knowledge of a \emph{probabilistically} safe action for each proxy state, together with an upper bound on the stopping time of the MDP. In our earlier work \cite{mazumdar2024safe}, the construction of a safe baseline policy relied on the assumption that a \emph{deterministically} safe action is available at every proxy state. In practice, however, such deterministically safe actions may not exist. This observation motivates the development of a safe baseline policy that accommodates a more general notion of safety, wherein safe actions are characterized in a probabilistic sense.

\begin{thm}[Safe baseline policy] \label{safe_pol}
Let $p\in(0,1)$ and suppose that for each state $x$ there exists a safe action $a^{s}(x)$ such that $h \leq \frac{p}{T_{\max}}$. Then, the policy $\pi^{S}$ defined below constitutes a safe baseline policy for all states $x \in H \cup U$:
\begin{equation*}
\pi^{S}(a \mid x) =
\begin{cases}
\check{\pi}^{S}(a \mid x'), & \text{if } x' \in U', \\[2mm]
\text{uniform random policy}, & \text{if } x' \in (H \setminus U') \cup U,
\end{cases}
\end{equation*}
where, for all $x' \in U'$, the policy $\check{\pi}^{S}$ is given by
\begin{equation*}
\check{\pi}^{S}(a \mid x') =
\begin{cases}
q, & \text{if } a = a^{s}(x'), \\[1mm]
\dfrac{1-q}{|\mathcal{A}|-1}, & \text{otherwise},
\end{cases}
\end{equation*}
with the mixing parameter $q$ satisfying $1 \geq q \geq \frac{1-\frac{p}{T_{\max}}}{1-h}$.
\end{thm}

\begin{pf}
We prove that the safety function for the initial state $x_0 \in H$ with the policy ${\pi}^S$ satisfies:
 \[
 S^P_{\pi^S}(x_0) \leq p.
 \]
Considering all the possible realizations, we obtain inequality $(a)$ as follows:
\begin{equation}
    \begin{split}
           & S^P_{\pi^S}(x_0)= \mathds{E}_{\pi^S}^{x_0} [\sum_{t=0}^{\tau-1} \kappa_t ; P] \\
           & \overset{(a)}{\leq} \mathds{P}_{\pi^S}^{x_0} [\tau\leq T_{max} ; P] \mathds{E}_{\pi^S}^{x_0} [\sum_{t=0}^{T_{max}-1} \kappa_t ; P]\\
           & + \mathds{P}_{\pi^S}^{x_0} [\tau> T_{max} ; P] \mathds{E}_{\pi^S}^{x_0} [\sum_{t=T_{max}}^{\infty} \kappa_t ; P]
           \end{split}
\end{equation}
Since $\mathds{P}_{\pi^S}^{x'} [\tau\leq T_{max} ; P]=1$, the above inequality becomes:
\[
S^P_{\pi^S}(x_0)\leq \mathds{E}_{\pi^S}^{x_0} [\sum_{t=0}^{T_{max}-1} \kappa_t ; P].
\]
Furthermore, for all $x\in H$,
\[
\mathds{E}_{\pi^s}^{x}\left[\kappa_t ; P \right]\leq \underset{x\in H}{\text{max}} \sum_{a\in A} \pi^S(a|x) \kappa(x,a), \ \forall t\in [0,T_{max}-1]
\]
Thus,
\[
\begin{split}
S^P_{\pi^S}(x_0) & \overset{}{\leq} T_{max} \ \underset{t\in [0,T_{max}-1]}{\text{max}} \ \mathds{E}_{\pi^S}^{x_0} \left[ \kappa_t ; P \right] \\
& \overset{}{\leq}  T_{max} \ \underset{{x}\in H}{\text{max}} \sum_{a\in A} \pi^S(a|x) \kappa(x,a) \\
& = T_{max} \ \underset{{x}\in H}{\text{max}} \sum_{y\in U} \Big[ q \cdot P(x,a^{s}(x),y) \\ 
            & + \sum_{a\in A\setminus \{a^{s}(x)\}}  \frac{(1-q)}{|\mathcal{A}|-1} \cdot P(x,a,y) \Big] \\
           & {=} T_{max} \ \Big[ \underset{x\in H}{\text{max}} \ q \ \sum_{y\in U}  P(x,a^{s}(x),y) \\
           &+ \underset{x\in H}{\text{max}}  \sum_{a\in A\setminus \{a^{s}(x)\}} \frac{(1-q)}{|\mathcal{A}|-1}  \sum_{y\in U} P(x,a,y) \Big ].
\end{split}
\]

Using the hypotheses, we obtain that the RHS is bounded as follows:
\[
\begin{split}
& \overset{}{\leq} T_{max} \Big(  q \ h + \frac{(|\mathcal{A}|-1)}{(|\mathcal{A}|-1)} (1-q) \Big) \\
           & = T_{max} \Big( 1 - q \ (1-h)) \Big) \\
           & \overset{(a)}{\leq} p.
\end{split}
\]
Inequality $(a)$ follows from the definition of the safe action. \qed
\end{pf}

The following result establishes the safety guarantee of the pSRL algorithm. The proof proceeds along similar lines to the arguments used in the proof of Theorem~2 in \cite{mazumdar2024safe}.

\begin{thm}[pSRL algorithm is $p$-safe] \label{Safety_proof}
Consider $\delta \in (0,\frac{1}{2})$. With probability at least $(1-2\delta)$, the policy ${\pi}_k$ designed using the pSRL algorithm is safe, i.e., the safety function $S^P_{{\pi}_k}({x}_0)$ for the initial state $x_0$ satisfies the following:
\begin{equation*}
    S^P_{{\pi}_k}({x}_0) \leq p.
\end{equation*} 
\end{thm}  
To evaluate the performance of RL algorithms, the notion of \textit{cumulative objective regret} or simply \textit{objective regret} $R(K)$ is commonly used.
\begin{defn}[Regret]
We define the per-episode objective regret and per-episode constraint regret as:
\begin{equation*}
\begin{split}
    &\mathcal{R}_k :=  V^P_{\pi_k}(c,x_0) - V^P_{\pi^*}(c,x_0),\\
    &\mathcal{C}_k :=  S^P_{\pi_k}(x_0) - S^P_{\pi^*}(x_0).
\end{split}
\end{equation*}
 The cumulative objective regret over $K$ episodes is defined as
\begin{equation*}
    R(K) := \sum_{k=1}^{K}\mathcal{R}_k.
\end{equation*}
\end{defn}
 \begin{thm}[\small{Regret bound of the pSRL algorithm}]\label{thm_regret_p_safe_RL}
Let $\delta \in \bigl(0,\tfrac{1}{2}\bigr)$. Then, with probability at least $(1-4\delta)$, the cumulative objective regret of the pSRL algorithm after $K$ episodes satisfies
\begin{equation*}
    R(K) \;\tilde{\mathcal{O}}\!\left(\frac{1}{p-p^{s}}\,|\mathcal{X}| \sqrt{|\mathcal{A}|\, T_{\max}^{3}\, K}\right),
\end{equation*}
where, $p$ denotes the safety threshold, $p^{s} := S^{P}_{\pi^{s}}(x_{0})$ is the safety level of the safe baseline policy $\pi^{s}$, $|\mathcal{X}|$ and $|\mathcal{A}|$ denote the numbers of states and actions, respectively, $T_{\max}$ is an upper bound on the stopping time, and $K$ is the number of episodes.
\end{thm}

 \section{Entropy regularized $p$-Safe RL (ER-pSRL) Algorithm} \label{entropy_algo}

 In this section, we present our main algorithm, \emph{entropy-regularized $p$-safe reinforcement learning (ER-pSRL)}, which enhances the optimization problem underlying the pSRL algorithm by incorporating an entropy regularization term.


\subsection*{Enhancing exploration and improving policy stability:}

The pSRL algorithm suffers from two serious issues: 
\begin{itemize}
\item During each episode $k$, a linear program is solved with modified cost coefficients given by
\[
\tilde{c}_k(x,a) = c(x,a) - \frac{4T_{\max}}{p - p^s} \hat{\epsilon}_k(x,a),
\]
where $\hat{\epsilon}_k(x,a)$ assigns higher values to less frequently visited state-action pairs $(x,a)$. Consequently, $\tilde{c}_k(x,a)$ is reduced more for such pairs, encouraging exploration. However, this method is somewhat heuristic, and a more principled approach to exploration can improve policy learning. 
\item With a linear objective, the optimum of an LP is attained at an extreme point of the feasible region. Consequently, even small perturbations in $\hat{P}_k$ can change the active constraint set and move the optimizer to a different vertex. As a result, the solution $(\hat h_k,\hat g_k)$ may vary significantly in response to minor changes in $(\hat{P}_k,\tilde c_k)$, leading to unstable policy updates and large episode-to-episode fluctuations in the per-episode objective regret, even when the underlying estimation noise is small.

\end{itemize}


To address the above issues, we introduce an entropy regularizer directly into the objective function of the optimization problem~\eqref{ofu_ext_lp}. Specifically, we add an entropy term 
that penalizes the negative entropy of the decision variables, encouraging more spread-out solutions. Since entropy is maximized when more decision variables are non-zero, the solution to the entropy-regularized optimization~\eqref{opt_entr_new} is typically less sparse. As a result, the induced policy, derived via~\eqref{estimate_pol_entr}, tends to be more exploratory. 

Further, the entropy regularizer induces strong convexity in the optimization problem, ensuring that the optimal occupation measures depend continuously on perturbations in both the empirical transition model $\hat{P}_k$. As a result, successive solutions vary smoothly across episodes, yielding stable policy updates and significantly reduced variability in the per-episode regret. 

 \par The ER-pSRL algorithm works almost exactly the same way as the pSRL algorithm. In each episode $k\in \{1,2,...,K\}$, instead of computing the policy from Extended LP \eqref{ofu_ext_lp}-\eqref{ofu_ext_lp_constraints}, we solve the following convex optimization.
\scriptsize
\begin{equation} \label{opt_entr_new}
   \begin{split}
        &\tilde{V}^{\tilde{P}_k}_{\pi_k}(c,x_0):= \underset{\hat{h}\geq 0, \hat{g}\geq 0}{\text{min}}  \sum_{(x,a)\in \mathcal{X}\times \mathcal{A}} \tilde{\xi}(x,a) {c}(x,a) - \alpha \hat{\epsilon}_k(x,a) H\big( \tilde{\xi}(x,a) \big) \\
        & \textbf{s.t. } \eqref{ofu_ext_lp_constraints}, 
   \end{split} 
\end{equation}
\normalsize
 \scriptsize
 \begin{equation*}
    \begin{split}
    & \text{where, }\\
       & \tilde{\xi}(x,a) = \sum_{y\in \mathcal{X}} \tilde{h}(x, a,y) + \tilde{g}(x, a,y); \ \forall (x,a,y)\in \mathcal{X}\times \mathcal{A} \times \mathcal{X}, \\
       & \eta \in (0,1),\\
       & \alpha = \frac{8 T_{max}^2}{\text{log} \left(\frac{2T_{max}}{T_{max}+\eta}  \right) (p-p^s)}, \\
     & H(\tilde{\xi}(x,a)) = - \frac{\tilde{\xi}(x,a)}{2T_{max}}  \ \text{log} \left(\frac{\tilde{\xi}(x,a)+\eta}{2T_{max}}  \right), \\
    \end{split}
 \end{equation*}
 \normalsize

 \par The policy at episode $k$ is given by 
 \scriptsize
\begin{equation}
 {\pi}_k(a|x)=\frac{\tilde{\xi}^*(x,a)}{\sum_{a'\in \mathcal{A}}\tilde{\xi}^*(x,a')},
 \label{estimate_pol_entr}
\end{equation} 
\normalsize
where $\tilde{\xi}^*$ is the solution of optimization \eqref{opt_entr_new}.

We introduce the entropy function $H$ based on the following considerations. From the definition of the occupation measures $\gamma(x,a)$ and $\beta(x,a)$, it follows that $\tilde{\xi}(x,a) \leq T_{\max}$ for all $(x,a) \in \mathcal{X} \times \mathcal{A}$. Accordingly, we normalize the occupation measure by $T_{\max}$ and define the entropy in terms of $\tilde{\xi}(x,a)/T_{\max}$ rather than $\tilde{\xi}(x,a)$, which ensures that $H \geq 0$ for all feasible solutions $\tilde{\xi}$. Furthermore, we introduce a small parameter $\eta \in (0,1)$, chosen to be sufficiently smaller than typical values of $\tilde{\xi}$. While the inclusion of $\eta$ has a negligible effect on the entropy term itself, it plays a crucial technical role in the derivation of the desired regret bound.


The pseudo-code for the ER-pSRL algorithm is given in Algorithm $2$.

\small
\begin{algorithm} \label{Entropy_algorithm}
  \caption{: ER-pSRL Algorithm}
  \begin{algorithmic}[1]
    \State \textbf{Input:} Safety parameter $p\in (0,1)$, initial state $x_0$, confidence parameter $\delta \in (0,1)$, one safe action $a^{s}(x')$ for each proxy state, cost $c(x,a)$ for all $(x,a)\in \mathcal{X} \times \mathcal{A}$
    \State \textbf{Initialize:} Select $N_{1}(x,a)=0$, $N_{1}(x,a,y)=0$ for all $(x,a,y)\in H\times \mathcal{A}\times \mathcal{X}$
    \For  {Episodes ($k=1,2,...,K$)}
    \State Set $X_0 \leftarrow x_0$
    \State Solve the optimization \eqref{opt_entr_new} using the estimate $\hat{P}_k(x,a,y)$ from \eqref{computed_prob}
    \If{optimization \eqref{opt_entr_new} is feasible}
    \State Obtain the policy ${\pi}_{k}$ using \eqref{estimate_pol_entr}
    \Else
    \State Set ${\pi}_{k} \leftarrow \pi^s$ 
    \EndIf
  \For  { ($t=0,1,2,...$)}
    \If  { $X_t$ is not a terminal state}
    \State  Apply action $A_t$ according to the policy ${\pi}_{k}$
    \State Observe the next state $X_{t+1}$  
      \Else 
     \State Terminate the loop
    \EndIf
     \EndFor
     \EndFor
  \end{algorithmic}
\end{algorithm}
\normalsize

The set of feasible solutions of the optimization is exactly the same as the extended LP \eqref{ofu_ext_lp}-\eqref{ofu_ext_lp_constraints}. Thus, since the policy designed using the pSRL algorithm is safe with an arbitrarily high probability, so is the policy designed using the ER-pSRL algorithm. We formally present this result as follows.
\begin{thm}[The ER-pSRL algorithm is $p$-safe] \label{Safety_proof}
Let $\delta \in \bigl(0,\tfrac{1}{2}\bigr)$. Then, with probability at least $(1-2\delta)$, the policy $\pi_k$ computed by the ER-pSRL algorithm is safe, i.e., for the initial state $x_0$, the associated safety function satisfies
\begin{equation*}
    S^{P}_{\pi_k}(x_0) \leq p
\end{equation*}
with probability at least $(1-2\delta)$.
\end{thm}

The regret bound for the ER-pSRL algorithm is as given below.
\begin{thm}[\scriptsize{Regret bound of the ER-pSRL algorithm}]\label{regret_entr_algo}
Let $\delta \in \bigl(0,\tfrac{1}{2}\bigr)$. Then, with probability at least $(1-4\delta)$, the cumulative objective regret of the ER-pSRL algorithm after $K$ episodes satisfies
\small
\begin{equation*}
    R(K) \;\tilde{\mathcal{O}}\!\left(|\mathcal{X}| \sqrt{|\mathcal{A}|\, T_{\max}^{3}\, K}\right),
\end{equation*}
\normalsize
where, $|\mathcal{X}|$ and $|\mathcal{A}|$ denote the numbers of states and actions, respectively, $T_{\max}$ is an upper bound on the stopping time, and $K$ is the number of episodes.
\end{thm}

\begin{rem}
The ER-pSRL algorithm eliminates the explicit dependence of the regret bound on $(p - p^s)^{-1}$. Since $0 < p - p^s < 1$, the regret bound associated with the OFU-based pSRL algorithm is always no smaller than its entropy-regularized counterpart. In particular, when the safe baseline policy operates close to the safety threshold ($p^s \approx p$), the regret bound for pSRL deteriorates sharply and becomes vacuous as $p \to p^s$. In contrast, ER-pSRL preserves a stable regret bound of order $\tilde{\mathcal{O}}\!\left(|\mathcal{X}|\sqrt{|\mathcal{A}|T_{\max}^3 K}\right)$, thereby demonstrating the effectiveness of entropy regularization in mitigating the adverse effects of a vanishing safety margin while maintaining sublinear regret.

\end{rem}

\section{Simulation results} \label{simulation_section}
In this section, we compare the efficacy of the ER-pSRL algorithm with the pSRL algorithm with the following example.

We consider a Markov decision process (MDP) defined as follows. The state space consists of five states,
$\mathcal{X}=\{1,2,3,4,5\}$, and the action space contains two actions,
$\mathcal{A}=\{1,2\}$. The living set is $H=\{1,2,3\}$, the unsafe state is
$U=\{4\}$, and the goal state is $E=\{5\}$. Both the unsafe and the goal states
are terminal. Clearly, the proxy state set is $\{2,3\}$. The nonzero state transition probabilities are given by
$P(1,1,2)=0.9,\; P(1,1,3)=0.1,\; P(1,2,2)=0.1,\; P(1,2,3)=0.9,\;
P(2,1,4)=0.8,\; P(2,1,5)=0.2,\; P(2,2,3)=0.2,\; P(2,2,5)=0.8,\;
P(3,1,4)=0.8,\; P(3,1,5)=0.2,\; P(3,2,5)=1.$ All remaining transition probabilities are zero. The safety threshold is set to $p=0.5$, and the confidence parameter is chosen as
$\delta = 0.01$. The upper bound on the
stopping time is chosen as $T_{\max}=5$, which almost surely upper bounds the
hitting time $\tau_E$. The safety function corresponding to the safe baseline
policy is computed as $p^{s}=S^{P}_{\pi^{s}}(1)=0.116$. The smoothing parameter is
set to $\eta=0.1$. The cost associated with each state-action pair is given
by $c(1,1)=-0.1$, $c(1,2)=-0.1$, $c(2,1)=-0.2$, $c(2,2)=-0.1$, $c(3,1)=-0.4$, and
$c(3,2)=-0.1$. The optimal policy is
$\pi^{*}(1|1)=0.46$, $\pi^{*}(2|1)=0.54$, $\pi^{*}(1|2)=0$,
$\pi^{*}(2|2)=1$, $\pi^{*}(1|3)=1$, and $\pi^{*}(2|3)=0$, yielding an
optimal value of $V^{P}_{\pi^{*}}=-0.396875$. We assume that a safe action is available at each proxy state $2$ and $3$, namely
action~$2$. Using Theorem~\ref{safe_pol}, we construct a safe baseline policy
$\pi^{s}$ given by $\pi^{s}(1|1)=\pi^{s}(2|1)=0.5$,
$\pi^{s}(1|2)=\pi^{s}(1|3)=0.1$, and
$\pi^{s}(2|2)=\pi^{s}(2|3)=0.9$. For state~$1$, a uniformly random policy is employed. 

\begin{figure}[!htb] 
  \includegraphics[scale=0.5]{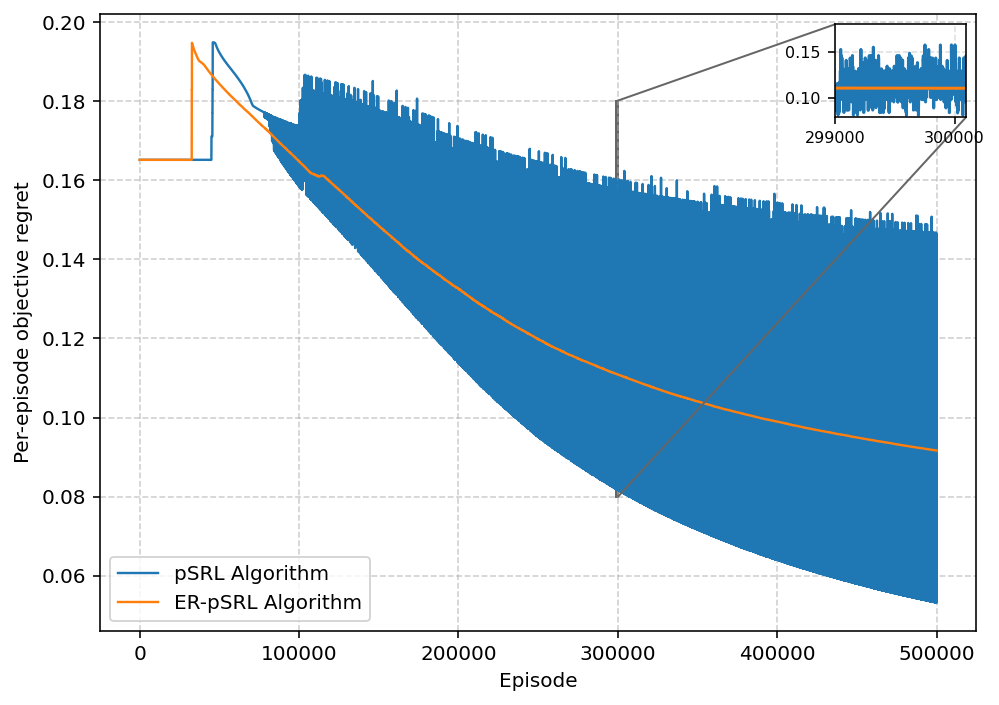}
\caption{Per-episode objective regret}
\label{fig:per_ep_reg}
\end{figure}

Figure~\ref{fig:per_ep_reg} illustrates the per-episode objective regret incurred
by the pSRL and ER-pSRL algorithms. It can be observed that the per-episode
objective regret of the pSRL algorithm exhibits significant variability across
episodes, indicating unstable policy updates. In contrast, the per-episode
objective regret of the ER-pSRL algorithm remains relatively stable, reflecting
more consistent policy improvement.

\begin{figure}[!htb] 
  \includegraphics[scale=0.5]{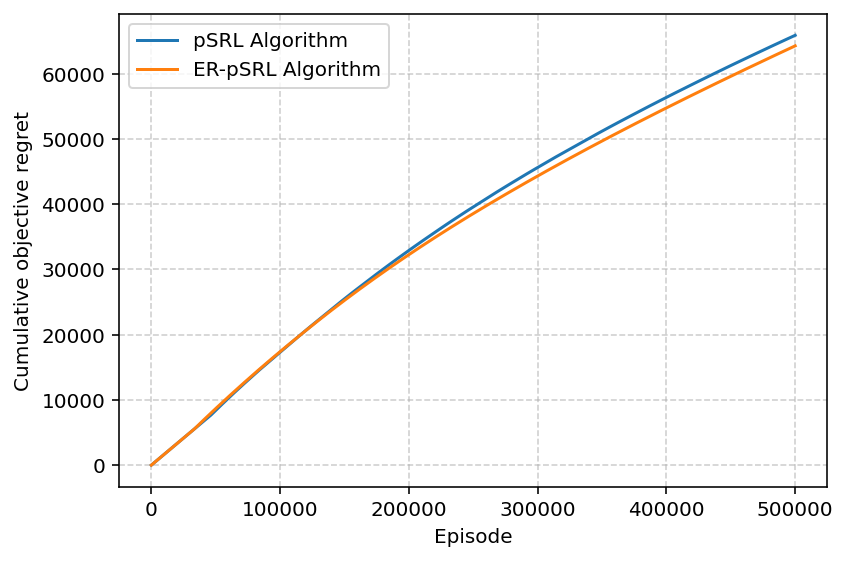}
\caption{Cumulative objective regret}
\label{fig:cum_reg_Algo_compar}
\end{figure}

\begin{figure}[!htb] 
  \includegraphics[scale=0.5]{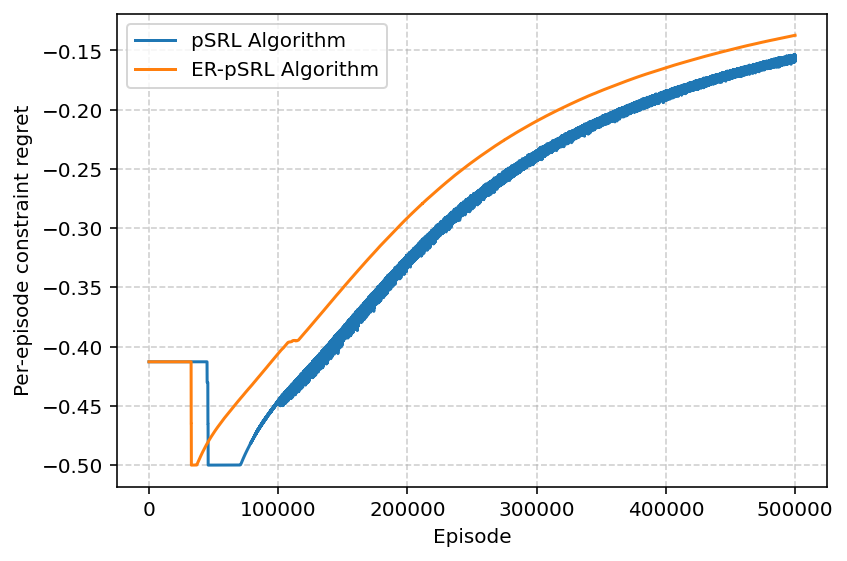}
\caption{Per-episode constraint regret}
\label{fig:per_ep_Cost_reg}
\end{figure}

Furthermore, the ER-pSRL algorithm demonstrates improved learning efficiency.
As shown in Figure~\ref{fig:cum_reg_Algo_compar}, its cumulative objective regret
remains consistently lower than that of the pSRL algorithm. Finally,
Figure~\ref{fig:per_ep_Cost_reg} presents the per-episode constraint regret for
both algorithms. For both pSRL and ER-pSRL, the per-episode constraint regret
remains below zero, indicating that the safety constraint is satisfied at all
times.



\begin{figure}[!htb] 
  \includegraphics[scale=0.47]{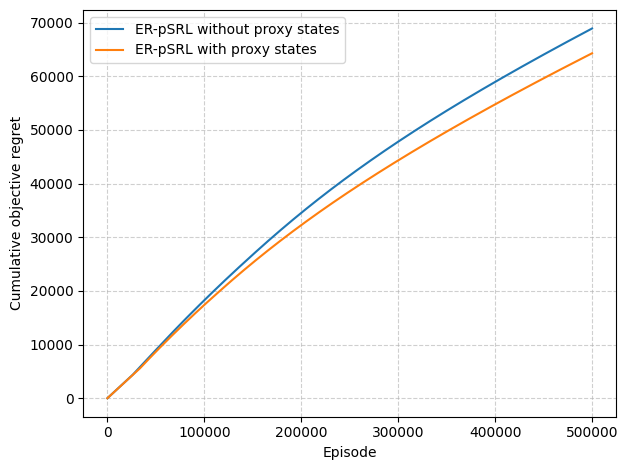}
\caption{Cumulative objective regret for the ER-pSRL algorithm}
\label{fig:rr_avg}
\end{figure}

We further compare the performance of the ER-pSRL algorithm under two scenarios:
\emph{(i)} when the proxy set $U'=\{2,3\}$ is known, and \emph{(ii)} when the proxy
set is unknown. When the proxy set is known, the safe baseline policy employed
until the optimization problem \eqref{opt_entr_new}--\eqref{ofu_ext_lp_constraints}
becomes feasible is the one described at the beginning of this section. In
contrast, when the proxy set $U'$ is unknown, we conservatively treat the entire
living set $H$ as the proxy set. In this case, the safe baseline policy is chosen
as $\pi^{s}(1\mid1)=\pi^{s}(1\mid2)=\pi^{s}(1\mid3)=0.1$ and
$\pi^{s}(2\mid1)=\pi^{s}(2\mid2)=\pi^{s}(2\mid3)=0.9$.

Figure~\ref{fig:rr_avg} shows that having prior knowledge of the proxy set leads
to a lower cumulative objective regret, indicating that a better policy is
learned. All simulation results correspond to averages taken over $5$ independent runs.

\section{Conclusion and Future Works} \label{sec_conclusion}
In this paper, we proposed two reinforcement learning algorithms that guarantee probabilistic safety with arbitrarily high confidence. We provided a finite-sample analysis of both algorithms and established explicit regret bounds. Our analysis revealed that the LP-based pSRL algorithm, while theoretically sound, suffers from significant episode-to-episode variability and insufficient exploration, which can degrade its empirical performance. To address these limitations, we introduced the ER-pSRL algorithm, which reduces high inter-episode variability and promotes systematic exploration in a principled manner. We theoretically demonstrated that ER-pSRL achieves a strictly improved regret bound compared to the LP-based pSRL algorithm.

As directions for future work, we plan to extend the ER-pSRL framework to settings with large or continuous state–action spaces by integrating function approximation techniques to enable generalization. Another promising avenue is to significantly accelerate the proposed entropy-regularized algorithm by leveraging ideas from the Sinkhorn algorithm. Since Sinkhorn-type methods efficiently solve entropically regularized linear programs arising in optimal transport, adapting these techniques could lead to scalable and computationally efficient implementations of ER-pSRL.

\bibliographystyle{IEEEtran}
\bibliography{abhi_cdc}

\section*{Appendix}
\addcontentsline{toc}{section}{Appendix}

 The rest of the proofs and results are based on a \textit{stopped Markov decision process} defined from the original MDP. 
\par Consider a \textit{stopped Markov decision process} ${M}^{\tau}$ with trajectory as $(X^{\tau}_t, A^{\tau}_t)$ defined as: \(X_t^{\tau} := X_{t \wedge \tau_E}, \quad A_t^{\tau} := A_{t \wedge (\tau_E-1)}, \quad \text{for all } t \geq 0\). The transition probability of the stopped MDP $M^{\tau}$ is as follows.  For all $ (x,a,y)\in (H\cup U)\times \mathcal{A} \times \mathcal{X}$, $P^{\tau}(x,a,y)=P(x,a,y)$ and for $ (x,a,y)\in E\times \mathcal{A} \times E$, $P^{\tau}(x,a,y)= 1$. The cost function $c$ defined for the original MDP is extended for the stopped MDP as follows for all $(X_t,A_t)\in \mathcal{X}\times \mathcal{A}$:

     \small
     \begin{equation*}
     c(X^{\tau}_t,A^{\tau}_t) = \begin{cases}
        c(X_t,A_t); \ \forall t\leq \tau_E-1\\
        0; \  \forall t \geq \tau_E
    \end{cases}
     \end{equation*}
     \normalsize
 We define the cumulative cost for the stopped MDP as follows:
     \scriptsize
\begin{equation*}
    V^{P^{\tau}}_{\pi}(c,x_0) := \mathds{E}^{{x}_0}_{\pi} \left[ \sum_{t=0}^{T_{max} -1} c(X^{\tau}_t,A^{\tau}_t) ; P^{\tau} \right]
\end{equation*}
\normalsize
     Suppose $X^\tau_{k,t}$ and $A^\tau_{k,t}$ represent the state and the action seen by the algorithm at time $t$ in episode $k$. For episode $k$, we define the following modified cost: 
      \small
     \begin{equation*}
     \tilde{c}_k(X^{\tau}_{k,t},A^{\tau}_{k,t}) = \begin{cases}
        \tilde{c}_k(X_{k,t},A_{k,t}); \ \forall t\leq \tau_E-1\\
        0; \  \forall t \geq \tau_E.
    \end{cases}
     \end{equation*}
     \normalsize
     For all time steps $t\geq \tau_E$ and all episodes $k$, the policy $\pi$ is defined as a deterministic policy, i.e., $\pi(X^{\tau}_{k,t})=a'$ where $a'$ is a any fixed action. 
     We also define $\sigma_{k,t}(x,a)$, the occupation measure for stopped MDP ${M}^{\tau}$ as follows:
     \scriptsize
     \begin{equation}
    \sigma_{k,t}(x,a):=  \mathds{P}_{\pi_k}[X^{\tau}_{k,t}=x,A_{k,t}^{\tau}=a ;  P^\tau | {X}^{\tau}_0={x}_0].
     \label{sigma_occupation}
     \end{equation}
     \normalsize
     The occupation measure $\sigma_{k,t}(x,a)$ satisfies the following property for all $t\geq 1$:
     \scriptsize
     \begin{equation}
     \begin{split}
    \sum_{a} \sigma_{k,t}(x,a)= \sum_{x',a'} P^\tau (x',a',x) \sigma_{k,t-1}(x',a')   
    \end{split}
    \label{occupation_property}
     \end{equation}
     \normalsize
For stopped MDP ${M}^{\tau}$, we use $\mathcal{F}^{\tau}_{k}$ to denote the filtration up to the the episode $k$ consisting of all state-action pairs visited until the end of episode $k$, i.e., $\mathcal{F}^{\tau}_k:\{X^\tau_{k',t}, A^\tau_{k',t}, k'\in [1,k] \ \text{and} \ t\leq T_{max}-1 \}$. We also define the following variables for each episode $k$: 
     \scriptsize
     \begin{equation*}
     \begin{split}
        & {N}^{\tau}_k(X^{\tau}_{k,t},A^{\tau}_{k,t}): = \begin{cases}
           & N_k(X_{k,t},A_{k,t}); \ \forall t\leq \tau_E -1 \\
            & 1; \ \forall t\geq \tau_E,
        \end{cases} 
    \end{split}
     \end{equation*}
\begin{equation}\label{eps_stop}
    \begin{split}
    & \hat{P}^{\tau}_k(X^{\tau}_{k,t},A^{\tau}_{k,t},X^{\tau}_{k,t+1}) := \begin{cases}
        & \hat{P}_k(X_{k,t},A_{k,t},X_{k,t+1}); \ \forall t\leq \tau_E-1 \\
        & P^{\tau}(X_{k,t},A_{k,t},X_{k,t+1}); \ \forall t\geq \tau_E,
    \end{cases} \\
       & \epsilon^{\tau}_k(x,a,y)\\
       &= \sqrt{\frac{4\hat{P}^{\tau}_k(x,a,y)(1-\hat{P}^{\tau}_k(x,a,y))L}{N^{\tau}_k(x,a)\vee 1}}  + \frac{14L}{3(N^{\tau}_k(x,a)\vee 1)}.  
    \end{split}
\end{equation}
 \normalsize

Since, target set $E$ is terminal so from the construction of the stopped MDP $M^{\tau}$, ${N}^{\tau}_k(x,a)=N_k(x,a)$ for all $k\in [1,K]$, $a\in \mathcal{A}$ and $x\in H\cup U$. We define the \emph{optimistic transition probability} $\tilde{P}^{\tau}_k$ for the stopped MDP $M^\tau$  as follows: 
     \scriptsize
     \begin{equation}
         \tilde{P}^{\tau}_k(X^{\tau}_{k,t},A^{\tau}_{k,t},X^{\tau}_{k,t+1}) := \begin{cases}
             & \tilde{P}_k(X_{k,t},A_{k,t},X_{k,t+1}); \ \forall t \leq \tau_E-1 \\
             & P^{\tau}(X_{k,t},A_{k,t},X_{k,t+1}); \ \forall t > \tau_E-1
         \end{cases}
         \label{prob_stop}
     \end{equation}
     \normalsize
     Note that if the transition probability $P$ of the original MDP is time invariant, so is $P^{\tau}$ for the stopped process $M^{\tau}$.
      \begin{lem}\label{concen_lem}
    With probability at least $(1-2\delta)$, the following is true for all $k\in \{1,2,...,K\}$ and for all $(x,a,y)\in \mathcal{X} \times \mathcal{A}\times \mathcal{\mathcal{X}}$:
    \begin{enumerate}[i)]
    \item   
 $|P^{\tau}(x,a,y) - \hat{P}^{\tau}_k(x,a,y)|\leq \epsilon^{\tau}_k(x,a,y)$.
 \normalsize 
 \\
    \item 
   $ |P^{\tau}(x,a,y)-\tilde{P}^{\tau}_k(x,a,y)|\leq 2\epsilon^{\tau}_k(x,a,y).$
    \normalsize
    \\
    \item 
    $ |\kappa(x,a) - \hat{\kappa}_k(x,a)| \leq  \hat{\epsilon}^{\tau}_k(x,a),$ \\
    where $\hat{\epsilon}^{\tau}_k(x,a)= \sum_{y\in \mathcal{X}} \epsilon^{\tau}_k(x,a,y).$
    \normalsize
    \end{enumerate} 
 \end{lem}
 \begin{pf}
     For all $(x,a,y)\in E\times \mathcal{A} \times \mathcal{X}$, claim $(i)$ and $(ii)$ are true from the definition. For all $(x,a,y)\in (H\cup U)\times \mathcal{A} \times \mathcal{X}$, claim $(i)$, $(ii)$ and $(iii)$ are proved in \cite{mazumdar2024safe}. \qed
 \end{pf}
\begin{lem} [Optimism of pSRL] \label{lem_optimism} 
    Suppose OFU \eqref{ofu_pol}-\eqref{ofu_pol_feas} is feasible. Then, the learned policy ${\pi}_k$ is an optimistic policy for the original MDP $M$, meaning that ${V}^{\tilde{P}_k}_{{\pi}_k}(\tilde{c}_k,x_0) \leq {V}^P_{{\pi}^*}({c},x_0)$. Further for the stopped MDP $M^{\tau}$,  ${V}^{\tilde{P}^{\tau}_k}_{{\pi}_k}(\tilde{c}_k,x_0) \leq {V}^{P^{\tau}}_{{\pi}^*}({c},x_0)$. 
 \end{lem}
\begin{pf}
    Consider the following optimization 
    \scriptsize
    \begin{equation}
        \begin{split}
         & \underset{\check{\pi}, \check{P} \in \mathcal{M}_k}{\text{min}} \ {V}^{\tilde{P}}_{\check{\pi}}(\check{c}_k,x_0) \\
         & \text{s.t. } \tilde{S}_{\check{\pi}}^{\check{P}}(x_0) \leq p
        \end{split}
        \label{aux_ofu}
    \end{equation}
    \normalsize
where
\scriptsize
\[\check{c}_k(x,a):= c(x,a) - 4w \hat{\epsilon}_k(x,a),\] \[\tilde{S}_{\check{\pi}}^{\check{P}}(x_0) := \mathds{E}_{\check{\pi}}^{x,\check{P}} \sum_{t = 0}^{\tau-1} \big(\hat{\kappa}(X_t, A_t) + 3 \hat{\epsilon}_k(X_t,A_t)\big),\] 
\normalsize
$\mathcal{M}_k$ is as defined in \eqref{para_ofu}.
\par Note that if $w = \frac{T_{max}}{p-p^s}$, then $\check{c}_k(x,a)= \tilde{c}_k(x,a)$, $\forall (x,a)\in \mathcal{X}\times \mathcal{A}$. Hence, optimization \eqref{aux_ofu} becomes equivalent to \eqref{ofu_pol}-\eqref{ofu_pol_feas} and ${V}^{\check{P}_k}_{\check{\pi}_k}(\check{c}_k,x_0)={V}^{\tilde{P}_k}_{\pi_k}(\tilde{c}_k,x_0)$, where $(\check{\pi}_k, \check{P}_k)$ is the optimal solution of \eqref{aux_ofu} and $(\pi_k,\tilde{P}_k)$ is the optimal solution of \eqref{ofu_pol}-\eqref{ofu_pol_feas}. In the sequel, we will show that if $w\geq \frac{T_{max}}{p-p_s}$, ${V}^{\check{P}_k}_{\check{\pi}_k}(\check{c}_k,x_0)\leq V^{{P}}_{\pi^*}(c,x_0)$.

\par With the true transition probability kernel $P$, suppose $(\gamma^s(x,a)$, $\beta^s(x,a)$) and $(\gamma^*(x,a)$, $\beta^*(x,a)$) be the occupation measure with respect to the safe baseline policy $\pi^s$ and the optimal policy $\pi^*$, respectively. Consider another occupation measure $(\gamma^m(x,a)$, $\beta^m(x,a)$) given by: $\gamma^m(x,a)=(1-\lambda)\gamma^s(x,a) + \lambda \gamma^*(x,a)$ and $\beta^m(x,a)=(1-\lambda)\beta^s(x,a) + \lambda \beta^*(x,a)$, where $\lambda \in [0,1]$. Suppose, $\pi^m$ is the policy corresponding to the occupation measures $(\gamma^m(x,a)$, $\beta^m(x,a)$). We derive a condition such that $(\pi^m, P)$ is a feasible solution of \eqref{aux_ofu}. Since the safety function ${S}^{{P}}_{\pi}(x_0)$ is a linear function of the occupation measure $\gamma(x,a)$, hence

\scriptsize
    \begin{equation}
        \begin{split}
            & \tilde{S}^{{P}}_{\pi^m}(x_0)  \\
            & = (1-\lambda) \tilde{S}^{{P}}_{\pi^s}(x_0) + \lambda \tilde{S}^{{P}}_{\pi^*}(x_0) \\
            & \overset{(a)}{{=}} (1-\lambda) \mathds{E}_{\pi^{s}}^{x_0} \left[ \sum_{t = 0}^{\tau-1} \big(\hat{\kappa}(X_t, A_t) + 3 \epsilon_k(X_t,A_t)\big) ; P \right] \\
            & + \lambda \mathds{E}_{\pi^{*}}^{x_0} \left[ \sum_{t = 0}^{\tau-1} \big(\hat{\kappa}(X_t, A_t) + 3 \epsilon_k(X_t,A_t)\big) ; P \right] \\
            & \overset{(b)}{{\leq}} (1-\lambda) \mathds{E}_{\pi^{s}}^{x_0} \left[ \sum_{t = 0}^{\tau-1} \big({\kappa}(X_t, A_t) + 4 \epsilon_k(X_t,A_t)\big) ; P \right] \\
            & + \lambda \mathds{E}_{\pi^{*}}^{x_0} \left[  \sum_{t = 0}^{\tau-1} \big({\kappa}(X_t, A_t) + 4 \epsilon_k(X_t,A_t)\big) ; P \right] \\
            & \overset{(c)}{{\leq}} (1-\lambda) \big({S}^{{P}}_{\pi^s}(x_0) + 4 \theta_k^{\pi_s} \big) + \lambda \big({S}^{{P}}_{\pi^*}(x_0) + 4 \theta_k^{\pi^*} \big) \\
            & \overset{(d)}{{\leq}} (1-\lambda) \big({p}_s + 4 \theta_k^{\pi_s} \big) + \lambda \big({p} + 4 \theta_k^{\pi^*} \big).
        \end{split}
    \end{equation}
    \normalsize
   Relation $(a)$ is from the definition of $\tilde{S}^{{P}}_{\pi}(x_0)$ from \eqref{aux_ofu}, $(b)$ follows from Lemma \eqref{concen_lem}, $(c)$ follows from the definition of ${S}^{{P}}_{\pi}(x_0)$ and $\theta_k^{\pi}:= \mathds{E}_{\pi}^{x_0} \left[ \sum_{t = 0}^{\tau_E-1} \hat{\epsilon}_k(X_t,A_t)  ; {P} \right]$, $(d)$ is due to the fact that ${p}_s:={S}^{{P}}_{\pi^s}(x_0)$ and ${S}^{{P}}_{\pi^*}(x_0)\leq {p}$. \\
    Now $(\pi^m, P)$ will be feasible solution of \eqref{aux_ofu} if 
    \scriptsize
    \begin{equation}
    \begin{split}
        & (1-\lambda) \big({p}_s + 4 \theta_k^{\pi_s} \big) + \lambda \big({p} + 4 \theta_k^{\pi^*} \big) \leq {p} \\
        & \implies \lambda \leq \frac{p - p^s - 4\theta_k^{\pi^s}}{p - p^s  - 4\theta_k^{\pi^s}+ 4\theta_k^{\pi^*}}.
    \end{split}
    \label{l_condition}
    \end{equation}
    \normalsize   

    Hence, ${V}_{\check{\pi}_k}^{\check{P}_k}(\check{c}_k,x_0)\leq {V}^P_{{\pi}^m}(\check{c}_k,x_0)$ if $\lambda=\frac{p - p^s - 4\theta_k^{\pi^s}}{p - p^s  - 4\theta_k^{\pi^s}+ 4\theta_k^{\pi^*}}$. Thus, ${V}_{\check{\pi}_k}^{\check{P}_k}(\check{c}_k,x_0)\leq {V}^P_{{\pi}^*}({c},x_0)$ if  ${V}^P_{{\pi}^m}(\check{c}_k,x_0)\leq {V}^P_{{\pi}^*}({c},x_0)$. We now prove that if $w\geq \frac{T_{max}}{p-p_s}$ then ${V}^P_{{\pi}^m}(\check{c}_k,x_0)\leq {V}^P_{{\pi}^*}({c},x_0)$. \\
    \scriptsize
    \begin{equation*}
    \begin{split}
        & {V}^P_{{\pi}^m}(\check{c}_k,x_0) \leq {V}^P_{{\pi}^*}({c},x_0) \\
        & \overset{a}{\implies} (1-\lambda) {V}^P_{{\pi}^s}(\check{c}_k,x_0) + \lambda {V}^P_{{\pi}^*}(\check{c}_k,x_0) \leq {V}^P_{{\pi}^*}({c},x_0) \\
        & \overset{b}{\implies} (1-\lambda) \big( {V}^P_{{\pi}^s}({c},x_0) - 4 w \theta_k^{\pi^s} \big) + \lambda \big( {V}^P_{{\pi}^*}({c},x_0) - 4 w \theta_k^{\pi^*} \big) \\
        & \hspace{6cm} \leq {V}^P_{{\pi}^*}({c},x_0) \\
        & {\implies} w \big(4 \lambda \theta_k^{\pi^*} + 4 (1-\lambda) \theta_k^{\pi^s} \big) \geq (1-\lambda) \big( {V}^P_{{\pi}^s}({c},x_0) - {V}^P_{{\pi}^*}({c},x_0) \big) \\
        & {\implies} w \geq \frac{{V}^P_{{\pi}^s}({c},x_0) - {V}^P_{{\pi}^*}({c},x_0)}{4 \theta_k^{\pi^s} + 4 \frac{\lambda}{(1-\lambda)} \theta_k^{\pi^*} }. 
    \end{split}
    \end{equation*}
    \normalsize

    The relation $(a)$ follows as  ${V}^P_{{\pi}}({c},x_0)$ is a linear function of the occupation measure, $(b)$ follows from the definition of $\check{c}_k$ and $\theta_k^\pi$. If we choose $\lambda=\frac{p - p^s - 4\theta_k^{\pi^s}}{p - p^s  - 4\theta_k^{\pi^s}+ 4\theta_k^{\pi^*}}$, then $4 \theta_k^{\pi^s} + 4 \frac{\lambda}{(1-\lambda)} \theta_k^{\pi^*} = p - p^s$. Now, since ${V}^P_{{\pi}^s}({c},x_0) - {V}^P_{{\pi}^*}({c},x_0) \leq T_{max}$, we get that $w \geq  \frac{T_{max}}{p-p^s} \geq \frac{{V}_{{\pi}^s}({c},x_0) - {V}_{{\pi}^*}({c},x_0)}{4 \theta_k^{\pi^*} + 4 \frac{\lambda}{(1-\lambda)} \theta_k^{\pi^s}}$. So, if we choose $w= \frac{T_{max}}{p-p_s}$ then ${V}^{\check{P}_k}_{\check{\pi}_k}(\check{c}_k,x_0) \leq {V}^P_{{\pi}^m}(\check{c}_k,x_0) \leq {V}^P_{{\pi}^*}({c},x_0)$. Hence,  ${V}^{\tilde{P}_k}_{{\pi}_k}(\tilde{c}_k,x_0) \leq {V}^P_{{\pi}^*}({c},x_0)$. 
    \par From the definition of the $c(X^{\tau}_t,A^{\tau}_t)$ and $\tilde{c}_k(X^{\tau}_t,A^{\tau}_t)$, we have that ${V}^{\tilde{P}^{\tau}_k}_{{\pi}_k}(\tilde{c}_k,x_0)={V}^{\tilde{P}_k}_{{\pi}_k}(\tilde{c}_k,x_0)$ and ${V}^{{P^{\tau}}}_{{\pi}^{*}}({c},x_0)={V}^{{P}}_{{\pi}^{*}}({c},x_0)$. Thus, we finally have \[{V}^{\tilde{P}^{\tau}_k}_{{\pi}_k}(\tilde{c}_k,x_0) \leq {V}^{P^\tau_k}_{{\pi}^*}({c},x_0). \qed\]     
\end{pf}
\par We define the following events $\mathcal{E}$ and $\mathcal{L}$ as follows:
    \small
    \begin{equation*}
        \mathcal{E}:= \Big{\{} \mathcal{M}_k, \ \forall k \in [1,K] \Big{\}},
    \end{equation*}
    $\mathcal{M}_k$ is as defined in \eqref{para_ofu}.

and
 \scriptsize
    \begin{equation*}
        \begin{split}
            & \mathcal{L}:= \Big{\{} N^{\tau}_{k-1}(x,a) \geq \frac{1}{2} \sum_{j<k} \sum_{t=0}^{T_{max}-1} \sigma_{j,t}(x,a) \\
            & - T_{max} \text{log}\left(\frac{|\mathcal{X}||\mathcal{A}|T_{max}}{\delta}\right), \ \forall (x,a)\in \mathcal{X}\times \mathcal{A}, \forall k\in [1,K] \big) \Big{\}}, 
        \end{split}
    \end{equation*}
    \normalsize
    where $\sigma_{j,t}(x,a)$ is defined in Equation \eqref{sigma_occupation}.
The following result is in line with \cite{zanette2019tighter} (Appedix D.4.)
    \begin{lem}
       For a chosen $\delta\in (0,1)$, $\mathds{P} \big[\mathcal{L} \big] \geq 1-2\delta . $
    \end{lem}
     We call the event $\mathcal{G}:= \mathcal{E} \cap \mathcal{L}$ as the \textbf{\textit{good event}}. Using union bound we have the following result.
    \begin{lem}
        For a chosen $\delta\in (0,1)$, $\mathds{P} \big[\mathcal{G} \big] \geq 1-4\delta. $
    \end{lem}
     \par For stopped MDP ${M}^{\tau}$, the following two lemmas are in line with Lemma $(38)$ and $(39)$ in \cite{efroni2019tight}, respectively.
    \begin{lem}\label{hat_sqrt_N_lem}
        Within the event $\mathcal{L}$, the following is true:
        \scriptsize
        \begin{equation*}
        \begin{split}
           \sum_{k=1}^{K} \sum_{t=0}^{T_{max}-1} \mathds{E} \big[ \frac{1}{\sqrt{{N}^{\tau}_k({X}^{\tau}_{k,t},{A}^{\tau}_{k,t})}} \big| {\mathcal{F}}^{\tau}_{k-1} \big] \\
           \overset{\overset{}{<}}{\sim}  \tilde{\mathcal{O}} \big(\sqrt{|\mathcal{X}| |\mathcal{A}| T_{max} K} + |\mathcal{X}| |\mathcal{A}| T_{max} \big).
        \end{split}
        \end{equation*}
        \normalsize
    \end{lem}

    \begin{lem} \label{hat_N_lem}
        Within the event $\mathcal{L}$, the following is true:
        \scriptsize
        \begin{equation*}
          \sum_{k=1}^{K} \sum_{t=0}^{T_{max}-1} \mathds{E} \big[ \frac{1}{{N}^{\tau}_k({X}^{\tau}_{k,t},{A}^{\tau}_{k,t})} \big| {\mathcal{F}}^{\tau}_{k-1} \big] \overset{\overset{}{<}}{\sim}  \tilde{\mathcal{O}} \big( |\mathcal{X}| |\mathcal{A}| T_{max} \big). 
        \end{equation*}
        \normalsize
    \end{lem}
\par  The following result is based on Lemma \ref{hat_sqrt_N_lem} and Lemma \ref{hat_N_lem}.
\begin{lem} \label{sum_eps_bound}
  Within the event $\mathcal{L}$, the following is true:
  \scriptsize
\begin{equation*}
    \begin{split}
       & \sum_{k=1}^{K} \ \sum_{t,x,a} \sigma_{k,t}(x,a) \sum_{y\in \mathcal{X}} \epsilon^{\tau}_k(x,a,y) \\
       & \overset{\overset{}{<}}{\sim}  \tilde{\mathcal{O}} \left(  \sqrt{|\mathcal{X}|^2 |\mathcal{A}| T_{max} K} + |\mathcal{X}|^2 |\mathcal{A}| T_{max}  \right).
    \end{split}
\end{equation*}
  \normalsize
\end{lem}
\begin{pf}
    \scriptsize
    \begin{equation*}
    \begin{split}
    & \sum_{k=1}^{K} \ \sum_{t,x,a} \sigma_{k,t}(x,a) \sum_{y\in \mathcal{X}} \epsilon^{\tau}_k(x,a,y) \\
       & = \sum_{k=1}^{K} \ \mathbb{E}^{x_0}_{\pi_k} \ \left[ \sum_{t=0}^{T_{max}-1} \sum_{y\in \mathcal{X}} \epsilon^{\tau}_k(X^{\tau}_{k,t},A^{\tau}_{k,t},y) ; P^{\tau} \right] \\
       & \overset{(a)}{\leq} \sum_{k=1}^K \mathbb{E}^{x_0}_{\pi_k} \Big[
\sum_{t=0}^{T_{max}-1}  \Big( \sqrt{\frac{4L}{N^{\tau}_k(X^{\tau}_{k,t},A^{\tau}_{k,t})\vee 1}} \sum_{y\in \mathcal{X}} \sqrt{\hat{P}^{\tau}_k(X^{\tau}_{k,t},A^{\tau}_{k,t},y)} \\
& + \sum_{y\in \mathcal{X}} \frac{(14/3)L}{N^{\tau}_k(X^{\tau}_{k,t},A^{\tau}_{k,t})\vee 1} ; P^{\tau}  \Big) \Big] \\
& \overset{(b)}{\leq} \sum_{k=1}^K \mathbb{E}^{x_0}_{\pi_k} \Big[ 2 \sqrt{L}
\sum_{t=0}^{T_{max}-1}  \sqrt{\frac{1}{N^{\tau}_k(X^{\tau}_{k,t},A^{\tau}_{k,t})\vee 1}} \\
& \cdot \sqrt{\sum_{y\in \mathcal{X}} \hat{P}^{\tau}_k(X^{\tau}_{k,t},A^{\tau}_{k,t},y)} \sqrt{|\mathcal{X}|}   \\
& + (14/3)L 
\sum_{t=0}^{T_{max}-1} \frac{1}{N^{\tau}_k(X^{\tau}_{k,t},A^{\tau}_{k,t})\vee 1} |\mathcal{X}| ; P^{\tau}    \Big] \\
& \overset{\overset{}{<}}{\sim}  \tilde{\mathcal{O}} \left(  \sqrt{|\mathcal{X}|^2 |\mathcal{A}| |T_{max}| K} + |\mathcal{X}|^2 |\mathcal{A}| T_{max}  \right).
    \end{split}
\end{equation*}
\normalsize
Relation $(a)$ follows from the definition of $\epsilon^{\tau}_k(x,a,y)$, relation $(b)$ follows as
\scriptsize
\begin{equation} \label{cauch_schar}
\sum_{y\in \mathcal{X}} \sqrt{\hat{P}^{\tau}_k(X^{\tau}_{k,t},A^{\tau}_{k,t},y)} \leq \sqrt{|\mathcal{X}|} \sqrt{\sum_{y\in \mathcal{X}} \hat{P}^{\tau}_k(X^{\tau}_{k,t},A^{\tau}_{k,t},y)}    
\end{equation}
 \normalsize
due to the Cauchy-Schwarz inequality. The final inequality follows from Lemma \ref{hat_sqrt_N_lem} and Lemma \ref{hat_N_lem}. \qed
\end{pf}

\begin{lem}
The value function $V^{P^\tau}_{\pi}(c,x)$ can be expressed recursively as follows:
\scriptsize
\begin{equation*}
    \begin{split}
    & V^{P^\tau}_{\pi}(c,x) = \mathds{E}^{x}_{\pi} \left[c(x,a) + \sum_{y\in \mathcal{X}} P^{\tau}(x,a,y) V^{P^\tau}_{\pi}(c,y) \Big| a \sim \pi(x) \right].
    \end{split}
\end{equation*}
\normalsize
\end{lem}

\par  We can prove the following result similar to Lemma $E.15$ in \cite{dann2017unifying}.  
    \begin{lem}[Value difference lemma] \label{value_diff_lem}
     For any policy $\pi$ and any $x\in H\cup U$, we have the following relation:
     \scriptsize
     \begin{equation*}
         \begin{split}
              & V^{P^\tau}_{\pi}(c,x) - V^{\bar{P}^{\tau}}_{\pi}(\bar{c},x) \\
              &= \mathds{E}^{x}_{\pi} \sum_{t=0}^{T_{max}-1} \Big[\left(c(X^{\tau}_{t},A^{\tau}_{t})-\bar{c}(X^{\tau}_{t},A^{\tau}_{t})\right)  \\
             &+  \sum_{y\in \mathcal{X}} \big(P^{\tau}(X^{\tau}_{t},A^{\tau}_{t},y) - \bar{P}^{\tau}(X^{\tau}_{t},A^{\tau}_{t},y)\big) V^{{P}^\tau}_{\pi}({c},y); \bar{P}^{\tau}  \Big] \\
             & = \mathds{E}^{x}_{\pi} \sum_{t=0}^{T_{max}-1} \Big[\left(c(X^{\tau}_{t},A^{\tau}_{t})-\bar{c}(X^{\tau}_{t},A^{\tau}_{t})\right)  \\
             &+  \sum_{y\in \mathcal{X}} \big(P^{\tau}(X^{\tau}_{t},A^{\tau}_{t},y) - \bar{P}^{\tau}(X^{\tau}_{t},A^{\tau}_{t},y)\big) V^{\bar{P}^{\tau}}_{\pi}(\bar{c},y) ; {P}^{\tau}  \Big].
         \end{split}
     \end{equation*}
     \normalsize
    \end{lem}
\begin{lem} \label{val_diff_2}
    For any $x\in \mathcal{X}$, the following is true for the stopped MDP $M^{\tau}$ within the event $\mathcal{E}$:
    \small
    \begin{equation*}
    \begin{split}
        & V^{\tilde{P}_k^{\tau}}_{\pi_k}(\tilde{c}_k,x) \leq V^{P^{\tau}}_{\pi_k}(c,x). 
    \end{split}
    \end{equation*}
    \normalsize
    \end{lem}
\begin{pf}
        Using the value difference lemma \ref{value_diff_lem},
        \scriptsize
        \begin{equation*}
            \begin{split}
             &    V^{\tilde{P}_k^\tau}_{\pi_k}(\tilde{c}_k,x) - V^{P^\tau}_{\pi_k}(c,x) \\
             & \overset{(a)}{=} \mathds{E}^{x}_{\pi_k} \sum_{t=0}^{T_{max}-1} \big[\left(\tilde{c}_k(X^{\tau}_t,A^{\tau}_t)-{c}(X^{\tau}_t,A^{\tau}_t)\right)  \\
             & +  \sum_{y\in \mathcal{X}} \big(\tilde{P}^{\tau}_k(X^{\tau}_t,A^{\tau}_t,y) - {P}^{\tau}(X^{\tau}_t,A^{\tau}_t,y)\big) V^{{P}^\tau}_{\pi}({c},y) ; \tilde{P}^{\tau}_k  \big] \\
             & \overset{(b)}{\leq} \mathds{E}^{x}_{\pi_k} \sum_{t=0}^{T_{max}-1} \Big[\left({c}(X^{\tau}_t,A^{\tau}_t)-\frac{4T_{max}}{p-p^s}\hat{\epsilon}^{\tau}_k(X^{\tau}_t,A^{\tau}_t)-{c}(X^{\tau}_t,A^{\tau}_t)\right)  \\
             & +  \sum_{y\in \mathcal{X}} \epsilon^{\tau}_k(X^{\tau}_t,A^{\tau}_t,y) T_{max} ; \tilde{P}^{\tau}_k \Big] \\
             & \overset{}{\leq} \mathds{E}^{x}_{\pi_k} \sum_{t=0}^{T_{max}-1} \Big[-\frac{4T_{max}}{p-p^s}\hat{\epsilon}^{\tau}_k(X^{\tau}_t,A^{\tau}_t) +   T_{max} \hat{\epsilon}^{\tau}_k(X^{\tau}_t,A^{\tau}_t) ; \tilde{P}^{\tau}_k \Big] \\
             & \overset{(c)}{\leq} 0.
            \end{split}
        \end{equation*}
        \normalsize
        Relation $(a)$ follows from Lemma \ref{value_diff_lem}. Inequality $(b)$ follows from the definition of $\tilde{c}_k(x,a)$, Lemma \ref{concen_lem} and due to the fact that $V^{P^\tau}_{\pi}(c,x)\leq T_{max}$ for all $x\in \mathcal{X}$. Finally, we get inequality $(c)$ as $\frac{T_{max}}{p-p^s}\geq T_{max}$.  \qed
    \end{pf}
 The following result is in line with Lemma $33$ in \cite{efroni2020exploration}.
    \begin{lem} \label{val_diff_3}
      For the stopped MDP $M^{\tau}$, the following is true:
      \scriptsize
      \begin{equation*}
        \begin{split}
         & \sum_{t,x,a} \sigma_{k,t+1}(x,a) \left(V^{P^\tau}_{\pi_k}(c,x) - V^{\tilde{P}_k^\tau}_{\pi_k}(\tilde{c}_k,x) \right)   \\
         & \leq T_{max} \Big[ \left(V^{P^\tau}_{\pi_k}(c,x_0) - V^{\tilde{P}_k^\tau}_{\pi_k}(\tilde{c}_k,x_0) \right) \\
        & + \sum_{t,x,a} \sigma_{k,t}(x,a)   \Big|\left< \left(P^{\tau}-\tilde{P}^{\tau}_k \right)(X_t,A_t), \left(V^{\tilde{P}_k^\tau}_{\pi_k}(\tilde{c}_k) - V^{P^\tau}_{\pi_k}(c)  \right) \right>\Big| \Big].
        \end{split}
      \end{equation*}
      \normalsize
    \end{lem}
In the sequel, we show derive the regret bound for the pSRL algorithm. \\ 
\\
 \textbf{\textit{Proof of Theorem \ref{thm_regret_p_safe_RL}:}} Regret is given by
 \scriptsize
 \begin{equation}
    \begin{split}
        & R(K)\\
        & = \sum_{k=1}^K \ \big(V^P_{\pi_k}(c,x_0) - V^P_{\pi^*}(c,x_0)\big)\\
        &  \overset{(a)}{=} \sum_{k=1}^K \ \big(V^{P^\tau}_{\pi_k}(c,x_0) - V^{{P}^\tau}_{\pi^*}(c,x_0) \big) \\
        & = \sum_{k=1}^K \ \big(V^{P^\tau}_{\pi_k}(c,x_0) - V^{\tilde{P}_k^\tau}_{\pi_k}(\tilde{c}_k,x_0) + V^{\tilde{P}_k^\tau}_{\pi_k}(\tilde{c}_k,x_0) - V^{P^\tau}_{\pi^*}(c,x_0)\big) \\
      &  \overset{(b)}{\leq} \sum_{k=1}^K \ \big(V^{P^\tau}_{\pi_k}(c,x_0) - V^{\tilde{P}_k^\tau}_{\pi_k}(\tilde{c}_k,x_0) \big) \\
      &  \overset{(c)}{\leq}   \underbrace{ \sum_{k=1}^K \ \mathbb{E}^{x_0}_{\pi_k}
\sum_{t=0}^{T_{max}-1}  \left[ \left(c(X^{\tau}_{k,t},A^{\tau}_{k,t}) - \tilde{c}_k(X^{\tau}_{k,t},A^{\tau}_{k,t})\right) ; P^{\tau} \right]
}_{\text{(I)}} 
      \\
      & + \underbrace{ \sum_{k=1}^K \mathbb{E}^{x_0}_{\pi_k} \sum_{t=0}^{T_{max}-1} \left[ \left< \left| \big(P^{\tau} - \tilde{P}^{\tau}_k\big) (X^{\tau}_{k,t},A^{\tau}_{k,t})\right|, V^{\tilde{P}_k^\tau}_{\pi_k}(\tilde{c}_k) \right> ; P^{\tau}  \right] }_{\text{(II)}}.
    \end{split}
    \label{regret_bound}
 \end{equation}
 \normalsize
 Inequality $(a)$ follows from the definition of the stopped process $M^\tau$ and Assumption \ref{cost_assum}. Relation $(b)$ follows from Lemma \ref{lem_optimism}. Finally, we get inequality $(c)$ from Lemma \ref{value_diff_lem}.\\

 We now bound the terms $(I)$ and $(II)$ within the good event $\mathcal{G}$, i.e., with probability at least $(1-4\delta)$.

We bound the term $(I)$ as follows. 
 \scriptsize
 \begin{equation}
 \begin{split}
  (I) & = \sum_{k=1}^K \ \mathbb{E}^{x_0}_{\pi_k}
\sum_{t=0}^{T_{max}-1}  \left[ \left(c(X^{\tau}_{k,t},A^{\tau}_{k,t}) - \tilde{c}_k(X^{\tau}_{k,t},A^{\tau}_{k,t})\right) ; P^{\tau} \right] \\
& \leq \frac{4T_{max}}{p-p^s} \sum_{k=1}^K \mathbb{E}^{x_0}_{\pi_k} \left[
\sum_{t=0}^{T_{max}-1}  \hat{\epsilon}^{\tau}_k(X^{\tau}_{k,t},A^{\tau}_{k,t}) ; P^{\tau} \right]\\
& \overset{\overset{(a)}{<}}{\sim}  \tilde{\mathcal{O}} \left( \frac{1}{p-p^s} \left( |\mathcal{X}| \sqrt{ |\mathcal{A}| T^3_{max} K} + |\mathcal{X}|^2 |\mathcal{A}| T^2_{max} \right)  \right).
 \end{split}
 \label{bound_I}
 \end{equation}
 \normalsize
 Relation $(a)$ in the above expression is due to Lemma \ref{sum_eps_bound}.

Now we bound term $(II)$:
 \scriptsize
 \begin{equation}
 \begin{split}
     & (II)\\
     & = \sum_{k,t} \mathbb{E}^{x_0}_{\pi_k} \left[  \sum_{y\in \mathcal{X}} \big| \left(P^{\tau} - \tilde{P}^{\tau}_k\right) (X^{\tau}_{k,t},A^{\tau}_{k,t},y) \big|  V^{\tilde{P}_k^\tau}_{\pi_k}(\tilde{c}_k,y) \ ; P^{\tau}  \right] \\
     & \overset{}{=} \sum_{k,t,x,a} \sigma_{k,t}(x,a)  \sum_{y\in \mathcal{X}} \big| \left(P^{\tau} - \tilde{P}^{\tau}_k\right) (x,a,y) \big| V^{\tilde{P}_k^\tau}_{\pi_k}(\tilde{c}_k,y)    \\
     & = \underbrace{ \sum_{k,t,x,a} \sigma_{k,t}(x,a)   \sum_{y\in \mathcal{X}} \big| \left(P^{\tau} - \tilde{P}^{\tau}_k\right) (x,a,y) \big| V^{{P}^\tau}_{\pi_k}({c},y) }_{(A.1)} \\
     & + \underbrace{ \sum_{k,t,x,a} \sigma_{k,t}(x,a)   \left< \big| \left(P^{\tau} - \tilde{P}^{\tau}_k\right) (x,a) \big|, \big(V^{\tilde{P}_k^\tau}_{\pi_k}(\tilde{c}_k) - V^{{P}^\tau}_{\pi_k}({c})  \big)\right>}_{(A.2)}.
 \end{split}
 \label{termII_bound}
 \end{equation}
 \normalsize
The term $(A.1)$ can be bounded as:
 \scriptsize
 \begin{equation}
    \begin{split}
      (A.1) &=  \sum_{k,t,x,a} \sigma_{k,t}(x,a)   \sum_{y\in \mathcal{X}} \big| \left(P^{\tau} - \tilde{P}^{\tau}_k\right) (x,a,y) \big| V^{{P}^\tau}_{\pi_k}({c},y)     \\
      & \overset{(a)}{\leq} T_{max}  \sum_{k,t,x,a} \sigma_{k,t}(x,a)   \sum_{y\in \mathcal{X}} \hat{\epsilon}^{\tau}_k(X^{\tau}_{k,t},A^{\tau}_{k,t},y)  \\
      & \overset{\overset{(b)}{<}}{\sim} \tilde{\mathcal{O}} \left(  \sqrt{|\mathcal{X}|^2 |\mathcal{A}| T^3_{max} K} + |\mathcal{X}|^2 |\mathcal{A}| T^2_{max}  \right).
    \end{split}
    \label{A_1_bound}
 \end{equation}
 \normalsize
 Relation $(a)$ holds due to the relation between $P^{\tau}$ and $\tilde{P}^{\tau}_k$ from Lemma \ref{concen_lem}, and as $V^{{P}^\tau}_{\pi_k}({c},y)\leq T_{max}$. Relation $(b)$ follows from Lemma \ref{sum_eps_bound}.
\\ Term $(A.2)$ can be bounded as:
 \scriptsize
 \begin{equation}
    \begin{split}
        & (A.2)\\
        & = \sum_{k,t,x,a} \sigma_{k,t}(x,a) \Big|  \left<  \left(P^{\tau} - \tilde{P}^{\tau}_k\right) (x,a), \left(V^{\tilde{P}^{\tau}_k}_{\pi_k}(\tilde{c}_k) - V^{{P}^\tau}_{\pi_k}({c}) \right) \right> \Big| \\
        & \leq \underbrace{ 2 \sqrt{L} \sum_{k,t,x,a}  \frac{\sigma_{k,t}(x,a)}{\sqrt{N^{\tau}_k(x,a)\vee 1}} \left<  \sqrt{\tilde{P}^{\tau}_k(x,a)}, \Big| V^{\tilde{P}_k^{\tau}}_{\pi_k}(\tilde{c}_k) - V^{{P}^{\tau}}_{\pi_k}({c}) \Big|\right>}_{(B.1)} \\
        & + \underbrace{ \frac{14L}{3}  \sum_{k,t,x,a} \sigma_{k,t}(x,a) \sum_{y\in \mathcal{X}} \frac{\Big| \left(V^{\tilde{P}_k^{\tau}}_{\pi_k}(\tilde{c}_k,y) - V^{{P}^{\tau}}_{\pi_k}({c},y) \right)\Big|}{{N^{\tau}_k(x,a)\vee 1}} }_{(B.2)}.
    \end{split}
    \label{A2_1}
 \end{equation}
 \normalsize
 The last inequality holds from Lemma \ref{concen_lem} and \eqref{eps_stop}.\\
 Term $(B.2)$ can be bounded as:
 \scriptsize
 \begin{equation}
     \begin{split}
     & (B.2) \\
     & = \frac{14L}{3}  \sum_{k,t,x,a} \sigma_{k,t}(x,a) \sum_{y\in \mathcal{X}} \frac{\Big| \left(V^{\tilde{P}_k^{\tau}}_{\pi_k}(\tilde{c}_k,y) - V^{{P}^{\tau}}_{\pi_k}({c},y) \right) \Big|}{{N^{{\tau}}_k(x,a)\vee 1}}  \\
     & \overset{\overset{(a)}{<}}{\sim}   \sum_{k,t,x,a}  \frac{\sigma_{k,t}(x,a)}{{N^{{\tau}}_k(x,a)\vee 1}} \sum_{y\in \mathcal{X}} \Big| \Big(V^{\tilde{P}_k^{\tau}}_{\pi_k}({c},y) + \frac{T_{max}}{p-p^s} \Theta_k^{\pi_k}(\tilde{P}^{\tau}_k) \\
     & + V^{{P}^{\tau}}_{\pi_k}({c},y) \Big) \Big| \\
    & \overset{\overset{(b)}{<}}{\sim}  |\mathcal{X}| \left( 2T_{max} + \frac{T^2_{max}|\mathcal{X}|L}{p-p^s} \right) \sum_{k=1}^K \mathbb{E}^{x_0}_{\pi_k} \left[ \sum_{t=0}^{T_{max}-1} \frac{1}{{N^{{\tau}}_k(x,a)\vee 1}} ; P^{\tau} \right] \\
     & \overset{\overset{(c)}{<}}{\sim}  |\mathcal{X}| \left( 2T_{max} + \frac{T^2_{max}|\mathcal{X}|L}{p-p^s} \right) \tilde{\mathcal{O}} \left( |\mathcal{X}| |\mathcal{A}| T_{max} \right) \\
     & \overset{\overset{}{<}}{\sim}  \tilde{\mathcal{O}} \left( \frac{T^3_{max}|\mathcal{X}|^2 |\mathcal{A}|}{p-p^s} \right).
     \end{split}
     \label{A2_2}
 \end{equation}
 \normalsize
Relation $(a)$ is from the definition of $V^{\tilde{P}_k^{\tau}}_{\pi_k}(\tilde{c}_k,x)$ and $\Theta_k^{\pi_k}(\tilde{P}^{\tau}_k):= \mathds{E}_{\pi_k}^{y} \left[ \sum_{t = 0}^{T_{max}-1} \hat{\epsilon}^\tau_k(X^\tau_{k,t},A^\tau_{k,t})  ; \tilde{P}^{\tau}_k \right]$. Relation $(b)$ follows from the fact that $V_{\pi}^P(c,x)\leq T_{max}$ for all $\pi, P$ and $\Theta_k^{\pi_k}(\tilde{P}^{\tau}_k) \overset{{<}} {\sim}T_{max}|\mathcal{X}|L$. Relation $(c)$ is due to Lemma \ref{hat_N_lem}.

 Similarly, we bound term $(B.1)$:
\scriptsize
\begin{equation}
\begin{split}
 &    (B.1) \\
 &= 2 \sqrt{L} \sum_{k,t,x,a}  \frac{\sigma_{k,t}(x,a)}{\sqrt{N^{\tau}_k(x,a)\vee 1}} \left<  \sqrt{\tilde{P}^{\tau}_k(x,a)}, \Big| V^{\tilde{P}_k^{\tau}}_{\pi_k}(\tilde{c}_k) - V^{{P}^{\tau}}_{\pi_k}({c}) \Big|\right>\\
 & \overset{\overset{(a)}{<}}{\sim}  \sum_{k,t,x,a} \sigma_{k,t}(x,a)\frac{\sqrt{|\mathcal{X}|}\sqrt{\left<\tilde{P}^{{\tau}}_k(x,a),\left(V^{\tilde{P}_k^\tau}_{\pi_k}(\tilde{c}_k) - V^{{P}^\tau}_{\pi_k}({c}) \right)^2\right>}}{\sqrt{N^{{\tau}}_k(x,a)\vee 1}} \\
 & \overset{\overset{(b)}{<}}{\sim}  \sqrt{|\mathcal{X}|} \left( \sum_{t,k,x,a} \frac{\sigma_{k,t}(x,a)}{N^{{\tau}}_k(x,a)} \right)^{\frac{1}{2}} \\
 & \times \left( \sum_{k,t,x,a,y} \sigma_{k,t}(x,a) \tilde{P}^{{\tau}}_k(x,a,y)\left(V^{\tilde{P}_k^\tau}_{\pi_k}(\tilde{c}_k,y) - V^{{P}^\tau}_{\pi_k}({c},y) \right)^2 \right)^{\frac{1}{2}} \\
 & \overset{\overset{(c)}{<}}{\sim}  \tilde{\mathcal{O}}\left( \sqrt{|\mathcal{X}|^2 |\mathcal{A}| T_{max}}  \right)\\
 & \times \left( \sum_{k,t,y,a} \sigma_{k,t+1}(y,a) \left(V^{\tilde{P}_k^\tau}_{\pi_k}(\tilde{c}_k,y) - V^{{P}^\tau}_{\pi_k}({c},y) \right)^2 \right)^{\frac{1}{2}}.
\end{split}
\end{equation}
\normalsize
To get $(a)$ we use the Cauchy-Schwarz inequality \eqref{cauch_schar} and we suppress the poly-log term $2\sqrt{L}$. Relation $(b)$ holds again due to the Cauchy-Schwarz inequality. Inequality $(c)$ follows from Lemma \ref{hat_N_lem} and the property \eqref{occupation_property}.
 \\ Since from Lemma \ref{val_diff_2}, $ V^{{P}^\tau}_{\pi_k}({c},y) - V^{\tilde{P}_k^\tau}_{\pi_k}(\tilde{c}_k,y)  \geq 0$,
\scriptsize
\begin{equation*}
    \begin{split}
    & \left(V^{\tilde{P}_k^\tau}_{\pi_k}(\tilde{c}_k,y) - V^{{P}^\tau}_{\pi_k}({c},y)\right)^2 \\
    & \overset{\overset{(a)}{<}}{\sim}  \tilde{\mathcal{O}}\left(\frac{1}{p-p^s} T^2_{max} |\mathcal{X}| L \right) \left(V^{{P}^\tau}_{\pi_k}({c},y) - V^{\tilde{P}_k^\tau}_{\pi_k}(\tilde{c}_k,y)\right). 
    \end{split}
\end{equation*}
\normalsize
Relation $(a)$ holds as $\left(V^{{P}^\tau}_{\pi_k}({c},y) - V^{\tilde{P}_k^\tau}_{\pi_k}(\tilde{c}_k,y)\right) \leq \tilde{\mathcal{O}}\left(\frac{1}{p-p^s} T^2_{max} |\mathcal{X}| L \right)$.
\\ So,
\scriptsize
\begin{equation}
    \begin{split}
    &(B.1) \\
    & \overset{\overset{}{<}}{\sim}  \left(\sqrt{\frac{ |\mathcal{X}|^3 |\mathcal{A}|T^3_{max}  L}{p-p^s} } \right)\\
    & \times \left( \sum_{k,t,y,a} \sigma_{k,t+1}(y,a) \left( V^{{P}^\tau}_{\pi_k}({c},y) - V^{\tilde{P}_k^\tau}_{\pi_k}(\tilde{c}_k,y) \right) \right)^{\frac{1}{2}} \\
    & \overset{\overset{(a)}{<}}{\sim}  \left(\sqrt{\frac{ |\mathcal{X}|^3 |\mathcal{A}|T^4_{max}  L}{p-p^s} } \right) \Big[ \Big( \sum_k V^{P^\tau}_{\pi_k}(c,x_0) - V^{\tilde{P}_k^\tau}_{\pi_k}(\tilde{c}_k,x_0)  \Big) ^{\frac{1}{2}} + \\
    &  \Big( \sum_{k,t,x,a} \sigma_{k,t}(x,a)  \left| \left<  \left(P^{\tau} - \tilde{P}^{\tau}_k\right) (x,a), \left(V^{\tilde{P}_k^\tau}_{\pi_k}(\tilde{c}_k) - V^{{P}^\tau}_{\pi_k}({c})  \right)  \right> \right|  \Big)^{\frac{1}{2}} \Big]
    \end{split}
    \label{A2_3}
\end{equation}
\normalsize
Relation $(a)$ follows from Lemma \ref{val_diff_3} and using the fact that $\sqrt{a+b}\leq \sqrt{a}+\sqrt{b}$. \\
Combining \eqref{A2_1}, \eqref{A2_2} and \eqref{A2_3}:
\scriptsize
\begin{equation}
\begin{split}
    &  \sum_{k,t,x,a} \sigma_{k,t}(x,a)    \left|\left< \left(P^{\tau} - \tilde{P}^{\tau}_k\right) (x,a), \left(V^{\tilde{P}_k^\tau}_{\pi_k}(\tilde{c}_k) - V^{{P}^\tau}_{\pi_k}({c})  \right)\right> \right| \\
    & \overset{\overset{(a)}{<}}{\sim}   \frac{T^3_{max}|\mathcal{X}|^2 |\mathcal{A}|}{p-p^s}\\
    & + \sqrt{\frac{ |\mathcal{X}|^3 |\mathcal{A}|T^4_{max}  L}{p-p^s} } \Big[ \Big(  \sum_k \Big( V^{P^\tau}_{\pi_k}(c,x_0) - V^{\tilde{P}_k^\tau}_{\pi_k}(\tilde{c}_k,x_0) \Big)  \Big) ^{\frac{1}{2}} \\
     & +\Big( \sum_{k,t,x,a}  \sigma_{k,t}(x,a)    \left| \left< \left(P^{\tau} - \tilde{P}^{\tau}_k\right) (x,a), \Big(V^{\tilde{P}_k^\tau}_{\pi_k}(\tilde{c}_k) - V^{{P}^\tau}_{\pi_k}({c})  \Big) \right> \right|   \Big)^{\frac{1}{2}} \Big] 
\end{split}  
\label{A_2_ineq}
\end{equation}
\normalsize
Inequality \eqref{A_2_ineq} can be expressed as $0\leq X \leq a + b \sqrt{X}$, where,  
\scriptsize
\begin{equation*}
\begin{split}
  &    X= \sum_{k,t,x,a} \sigma_{k,t}(x,a)    \left|\left< \left(P^{\tau} - \tilde{P}^{\tau}_k\right) (x,a), \left(V^{\tilde{P}_k^\tau}_{\pi_k}(\tilde{c}_k) - V^{{P}^\tau}_{\pi_k}({c})  \right) \right> \right|, \\
 & a= \frac{T^3_{max}|\mathcal{X}|^2 |\mathcal{A}|}{p-p^s}\\
 &+ \sqrt{\frac{ |\mathcal{X}|^3 |\mathcal{A}|T^4_{max}  L}{p-p^s} } \Big(  \sum_k \Big( V^{P^\tau}_{\pi_k}(c,x_0) - V^{\tilde{P}_k^\tau}_{\pi_k}(\tilde{c}_k,x_0) \Big)  \Big) ^{\frac{1}{2}}, \\
 & b=\sqrt{\frac{ |\mathcal{X}|^3 |\mathcal{A}|T^4_{max}  L}{p-p^s} }.
\end{split}
\end{equation*}
\normalsize
Now if $0\leq X \leq a + b \sqrt{X}$, then we have that $X\leq a + b^2$ (Lemma 38 \cite{efroni2020exploration}). Thus,
\scriptsize
\begin{equation}
\begin{split}
    & (A.2) \\
     & = \sum_{k,t,x,a} \sigma_{k,t}(x,a)  \Big| \left< \left(P^{\tau} - \tilde{P}^{\tau}_k\right) (x,a), \left(V^{\tilde{P}_k^\tau}_{\pi_k}(\tilde{c}_k) - V^{{P}^\tau}_{\pi_k}({c})  \right)\right> \Big| \\
    & \overset{\overset{(a)}{<}}{\sim}  \frac{T^3_{max}|\mathcal{X}|^2 |\mathcal{A}|}{p-p^s} \\
    & + \sqrt{\frac{ |\mathcal{X}|^3 |\mathcal{A}|T^4_{max}  L}{p-p^s} } \Big(  \sum_k \Big( V^{P^\tau}_{\pi_k}(c,x_0) - V^{\tilde{P}_k^\tau}_{\pi_k}(\tilde{c}_k,x_0) \Big)  \Big) ^{\frac{1}{2}} \\
    & + \frac{ |\mathcal{X}|^3 |\mathcal{A}|T^4_{max}  L}{p-p^s}.
\end{split}  
\label{A_2_bound}
\end{equation}
\normalsize
Substituting \eqref{A_1_bound} and \eqref{A_2_bound} into \eqref{termII_bound}, we have the following bound for $(II)$.
\scriptsize
\begin{equation}
\begin{split}
& (II) \\
& \overset{\overset{}{<}}{\sim}  \sqrt{|\mathcal{X}|^2 |\mathcal{A}| T^3_{max} K} + |\mathcal{X}|^2 |\mathcal{A}| T^2_{max}+ \frac{T^3_{max}|\mathcal{X}|^2 |\mathcal{A}|}{p-p^s}  \\
&  + \sqrt{\frac{ |\mathcal{X}|^3 |\mathcal{A}|T^4_{max}  L}{p-p^s} } \Big(  \sum_k \Big( V^{P^\tau}_{\pi_k}(c,x_0) - V^{\tilde{P}_k^\tau}_{\pi_k}(\tilde{c}_k,x_0) \Big)  \Big) ^{\frac{1}{2}} \\
    & + \frac{ |\mathcal{X}|^3 |\mathcal{A}|T^4_{max}  L}{p-p^s}. 
\end{split}
\label{bound_II}
\end{equation}
\normalsize
Now from Inequalities \eqref{regret_bound}, \eqref{bound_I} and \eqref{bound_II}:
\scriptsize
\begin{equation}
\begin{split}
& \sum_k \Big( V^{P^\tau}_{\pi_k}(c,x_0) - V^{\tilde{P}_k^\tau}_{\pi_k}(\tilde{c}_k,x_0) \Big)  \\ 
& \overset{\overset{}{<}}{\sim}  \frac{|\mathcal{X}| \sqrt{ |\mathcal{A}| T^3_{max} K} }{p-p^s} + \frac{|\mathcal{X}|^2 |\mathcal{A}| T^2_{max}}{p-p^s} + \sqrt{|\mathcal{X}|^2 |\mathcal{A}| T^3_{max} K} \\
& + |\mathcal{X}|^2 |\mathcal{A}| T^2_{max}  + \frac{T^3_{max}|\mathcal{X}|^2 |\mathcal{A}|}{p-p^s} + \frac{ |\mathcal{X}|^3 |\mathcal{A}|T^4_{max}  L}{p-p^s} \\
& +\sqrt{\frac{ |\mathcal{X}|^3 |\mathcal{A}|T^4_{max}  L}{p-p^s} } \Big(  \sum_k \Big( V^{P^\tau}_{\pi_k}(c,x_0) - V^{\tilde{P}_k^\tau}_{\pi_k}(\tilde{c}_k,x_0) \Big)  \Big) ^{\frac{1}{2}} \\
& \overset{\overset{(a)}{<}}{\sim}   \sqrt{\frac{ |\mathcal{X}|^3 |\mathcal{A}|T^4_{max}  L}{p-p^s} } \Big(  \sum_k \Big( V^{P^\tau}_{\pi_k}(c,x_0) - V^{\tilde{P}_k^\tau}_{\pi_k}(\tilde{c}_k,x_0) \Big)  \Big) ^{\frac{1}{2}} \\
& + \frac{ \sqrt{ |\mathcal{X}|^2 |\mathcal{A}| T^3_{max} K} }{p-p^s}.
\end{split}
\end{equation}
\normalsize 
To get relation $(a)$ in the above expression, we ignore the terms that do not depend on episode $K$.

Now again consider $X=\sum_k \Big( V^{P^\tau}_{\pi_k}(c,x_0) - V^{\tilde{P}^\tau_k}_{\pi_k}(\tilde{c}_k,x_0) \Big) $, $a = \frac{ \sqrt{ |\mathcal{X}|^2 |\mathcal{A}| T^3_{max} K} }{p-p^s}$ and $b=\sqrt{\frac{ |\mathcal{X}|^3 |\mathcal{A}|T^4_{max}  L}{p-p^s} }$. Then using the fact that $0\leq X \leq a + b\sqrt{X}$ implies that $X \leq a + b^2$, we have the following inequality:
\scriptsize
\begin{equation}
\begin{split}
& \sum_k \Big( V^{P^\tau}_{\pi_k}(c,x_0) - V^{\tilde{P}_k^\tau}_{\pi_k}(\tilde{c}_k,x_0) \Big) \\
& \overset{\overset{}{<}}{\sim}  \frac{ \sqrt{ |\mathcal{X}|^2 |\mathcal{A}| T^3_{max} K} }{p-p^s} + \frac{ |\mathcal{X}|^3 |\mathcal{A}|T^4_{max}  L}{p-p^s}.
\end{split}
\end{equation}
\normalsize
Therefore, from inequality \eqref{regret_bound}, we have the following regret bound:
\scriptsize
\begin{equation*}
    \begin{split}
 R(K) & \overset{<}{\sim}  \frac{ \sqrt{ |\mathcal{X}|^2 |\mathcal{A}| T^3_{max} K} }{p-p^s} + \frac{ |\mathcal{X}|^3 |\mathcal{A}|T^4_{max}  L}{p-p^s} \\
& \overset{<}{\sim}  \tilde{\mathcal{O}}\left(\frac{1}{p-p^s}\sqrt{ |\mathcal{X}|^2 |\mathcal{A}| T^3_{max} K}  \right). \qed
\end{split}
\end{equation*}
\normalsize 

\begin{lem}\label{entr_ofu_cost_lem}
    Suppose, at episode $k$, $(\pi_k,\tilde{P}_k)$ is a solution of the OFU optimization \eqref{ofu_ext_lp}-\eqref{ofu_ext_lp_constraints}, then following is true:
    \begin{equation*}
        \tilde{V}^{\tilde{P}_k}_{\pi_k}(c,x_0) \leq V^{\tilde{P}_k}_{\pi_k}(\tilde{c}_k,x_0).
    \end{equation*}
\end{lem}
\begin{pf}
    From Optimization \eqref{opt_entr_new}-\eqref{ofu_ext_lp_constraints},
    \scriptsize
\begin{equation}  \label{Val_entr}
   \begin{split}
        &\tilde{V}^{\tilde{P}_k}_{\pi_k}(c,x_0)= V^{\tilde{P}_k}_{\pi_k}(c,x_0) - \alpha \sum_{(x,a)\in \mathcal{X}\times \mathcal{A}} \ \hat{\epsilon}_k(x,a)\ H\big( \tilde{\xi}^*(x,a) \big) \\
        & \overset{(a)}{\leq}  V^{\tilde{P}_k}_{\pi_k}(c,x_0)\\
        &- \frac{\alpha}{2T_{max}}   \text{log}\left(\frac{2T_{max}}{T_{max}+\eta} \right) \sum_{(x,a)\in \mathcal{X}\times \mathcal{A}} \tilde{\xi}^*(x,a) \hat{\epsilon}_k(x,a). 
   \end{split} 
\end{equation}
\normalsize
Relation $(a)$ is due to the fact that $H\big( \tilde{\xi}(x,a) \big)\geq    \text{log}\left(\frac{2T_{max}}{T_{max}+\eta} \right)\frac{\tilde{\xi}(x,a)}{2T_{max}} $, $\forall \xi$. We  can express $V^{\tilde{P}_k}_{\pi_k}(\tilde{c}_k,x_0)$ as:
\scriptsize
\begin{equation} \label{val_ofu}
    \begin{split}
      V^{\tilde{P}_k}_{\pi_k}(\tilde{c}_k,x_0)
     & \overset{}{=} V^{\tilde{P}_k}_{\pi_k}({c},x_0) - \frac{4T_{max}}{p-p^s}  \sum_{(x,a)\in \mathcal{X}\times \mathcal{A}} \tilde{\xi}^*(x,a) \hat{\epsilon}_k(x,a).
    \end{split}
\end{equation}
\normalsize
From \eqref{Val_entr} and \eqref{val_ofu}, if $\alpha = \frac{8T^2_{max}}{\text{log}(\frac{2T_{max}}{T_{max}+\eta})(p-p^s)} $, we get $\tilde{V}^{\tilde{P}_k}_{\pi_k}(c,x_0) \leq {V}^{\tilde{P}_k}_{\pi_k}(\tilde{c}_k,x_0)$. \qed
\end{pf}

\textbf{\textit{Proof of Theorem \ref{regret_entr_algo}:}} \\
\\
Suppose, at episode $k$, $\left(\pi_k^{1},\tilde{P}_k^{1}\right)$ and $\left(\pi_k^{2},\tilde{P}_k^{2}\right)$ are the solutions of the pSRL algorithm and the ER-pSRL algorithm, respectively. Since the constraints are the same for both algorithms, each of the solutions is feasible for both algorithms. So, we get that $\tilde{V}^{\tilde{P}_k^{2}}_{\pi^{2}_k}({c},x_0) \leq \tilde{V}^{\tilde{P}_k^{1}}_{\pi^{1}_k}({c},x_0)$. From Lemma \ref{entr_ofu_cost_lem}, $ \tilde{V}^{\tilde{P}_k^{1}}_{\pi^{1}_k}({c},x_0) \leq {V}^{\tilde{P}_k^{1}}_{\pi^{1}_k}(\tilde{c}_k,x_0)$.
From Lemma \ref{lem_optimism}, we have that ${V}^{\tilde{P}_k^{1}}_{\pi^{1}_k}(\tilde{c}_k,x_0)  \leq V^P_{\pi^*}(c,x_0)$. Hence,
\small
\begin{equation}\label{ineq_entr_ofu}
    \begin{split}
      & \tilde{V}^{\tilde{P}_k^{2}}_{\pi^{2}_k}({c},x_0) \leq {V}^{\tilde{P}_k^{1}}_{\pi^{1}_k}(\tilde{c}_k,x_0) \leq  V^P_{\pi^*}(c,x_0).
    \end{split}
\end{equation}
\normalsize
\scriptsize
\begin{equation*}
\begin{split}
     &R(K) \\
     &= \sum_{k=1}^{K} V_{\pi_k^{2}}^P(c,x_0) - V_{\pi^{*}}^P(c,x_0) \\
    & = \sum_{k=1}^{K} V_{\pi_k^{2}}^P(c,x_0) - \tilde{V}^{\tilde{P}_k^{2}}_{\pi^{2}_k}({c},x_0) + \tilde{V}^{\tilde{P}_k^{2}}_{\pi^{2}_k}({c},x_0) - V_{\pi^{*}}^P(c,x_0) \\
    & \overset{(a)}{\leq} \sum_{k=1}^{K} V_{\pi_k^{2}}^P(c,x_0) - \tilde{V}^{\tilde{P}_k^{2}}_{\pi^{2}_k}({c},x_0) \\
    & \overset{(b)}{\leq} \sum_{k=1}^{K} V_{\pi_k^{2}}^P(c,x_0) - {V}^{\tilde{P}_k^{2}}_{\pi^{2}_k}({c},x_0) + \alpha \sum_{x,a} \hat{\epsilon}_k(x,a) H(\tilde{\xi}^*(x,a)) \\
    & \overset{(c)}{\leq} \sum_{k=1}^{K} V_{\pi_k^{2}}^{P^\tau}(c,x_0) - {V}^{\tilde{P}_k^{2,\tau}}_{\pi^{2}_k}({c},x_0) \\
    & + \alpha \sum_{x,a} \text{log}\left(\frac{2T_{max}}{\eta} \right)\frac{\tilde{\xi}^*(x,a)}{2T_{max}}\hat{\epsilon}_k(x,a) \\
     &\overset{(d)}{=} \sum_{k=1}^K \mathbb{E}^{x_0}_{\pi_k} \sum_{t=0}^{T_{max}-1} \big[   \left< \big| \left(P^{\tau} - \tilde{P}^{2,\tau}_k\right) (X^{\tau}_{k,t},A^{\tau}_{k,t}) \big|, V^{\tilde{P}_k^{2,\tau}}_{\pi_k}({c})\right>; P^{2,\tau}   \big]   \\
    & + \frac{\alpha}{2T_{max}}   \text{log}\left(\frac{2T_{max}}{\eta} \right) \sum_{k,t,x,a} \sigma_{k,t}(x,a) \hat{\epsilon}^\tau_k(x,a) \\ 
    & \overset{<}{\sim} \tilde{\mathcal{O}}\left( |\mathcal{X}| \sqrt{|\mathcal{A}|T^3_{max}K} \right).
\end{split}
\end{equation*}
\normalsize
We get inequality $(a)$ in view of \eqref{ineq_entr_ofu}. Inequality $(b)$ is from the definition of $\tilde{V}^{\tilde{P}_k^2}_{\pi_k^2}(c,x_0)$. Relation $(c)$ follows from the fact that $H(\tilde{\xi}^*(x,a))\leq \text{log}\left(\frac{2T_{max}}{\eta} \right)\frac{\tilde{\xi}^*(x,a)}{2T_{max}}$. $ \tilde{P}^{2,\tau}_k$ is defined according to \eqref{prob_stop}. Relation $(d)$ is due to Lemma \ref{value_diff_lem}. To get the final relation, we use \eqref{A_1_bound} in the proof of Theorem \ref{thm_regret_p_safe_RL} and Lemma \ref{sum_eps_bound}. \qed

\end{document}